\documentclass{article}

\usepackage{arxiv}
\usepackage{comment}
\usepackage[utf8]{inputenc} 
\usepackage[T1]{fontenc}    
\usepackage{lineno}
\usepackage{hyperref}       
\hypersetup{hidelinks}
\usepackage{url}            
\usepackage{float,booktabs}       
\usepackage{amsfonts,amssymb,amsmath,bm,mleftright,relsize,siunitx,mathtools,amsthm,cancel,accents}
\usepackage{pifont}
\newcommand{\cmark}{\ding{51}}%
\newcommand{\xmark}{\ding{53}}%
\usepackage[bb=boondox]{mathalpha} 
\usepackage[normalem]{ulem}
\DeclareMathOperator*{\argmin}{arg\,min} 
 
\usepackage{scalerel} 
\modulolinenumbers[5]
\def\msquare{\mathord{\scalerel*{\Box}{gX}}}\usepackage{lipsum}         
\usepackage{graphicx}

\usepackage{nicefrac}       

\usepackage{microtype}      
\usepackage[capitalise]{cleveref}
\usepackage{graphicx}
\usepackage[numbers]{natbib}

\usepackage{makecell}
\usepackage{multirow}
\usepackage{algorithm,algpseudocode}
\usepackage{xcolor, soul}
\sethlcolor{green}

\usepackage{doi}
\usepackage{adjustbox}

\usepackage{arydshln}
\makeatletter
\def\adl@drawiv#1#2#3{%
        \hskip.5\tabcolsep
        \xleaders#3{#2.5\@tempdimb #1{1}#2.5\@tempdimb}%
                #2\z@ plus1fil minus1fil\relax
        \hskip.5\tabcolsep}
\newcommand{\cdashlinelr}[1]{%
  \noalign{\vskip\aboverulesep
           \global\let\@dashdrawstore\adl@draw
           \global\let\adl@draw\adl@drawiv}
  \cdashline{#1}
  \noalign{\global\let\adl@draw\@dashdrawstore
           \vskip\belowrulesep}}
\makeatother

\makeatletter

\newcommand{\volumedash}{%
  \makebox[0pt][l]{%
    \ooalign{\hfil\hphantom{$\m@th V$}\hfil\cr\kern0.08em--\hfil\cr}%
  }%
}
\makeatother
\makeatletter
\def\mathcolor#1#{\@mathcolor{#1}}
\def\@mathcolor#1#2#3{%
  \protect\leavevmode
  \begingroup\color#1{#2}#3\endgroup
}
\makeatother

\DeclareRobustCommand{\rchi}{{\mathpalette\irchi\relax}}
\newcommand{\irchi}[2]{\raisebox{\depth}{$#1\chi$}} 

\title{Exact identification of nonlinear dynamical systems by Trimmed Lasso}

\date{May 1, 2023}

\author{ \href{https://orcid.org/0000-0001-9202-0671}{\includegraphics[scale=0.06]{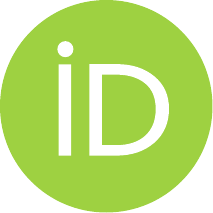}\hspace{1mm}Shawn L.~Kiser}\thanks{CNRS, CNAM, HESAM Université} \\
	Laboratoire PIMM\\
	Arts et Métiers Sciences et Technologies\\
	\texttt{shawn\_lee.kiser@ensam.eu} \\
\And
  \href{https://orcid.org/0000-0001-5877-9449}{\includegraphics[scale=0.06]{orcid.pdf}\hspace{1mm}}{Mikhail Guskov}$^*$ \\
    Laboratoire PIMM\\
	Arts et Métiers Sciences et Technologies\\
	\texttt{mikhail.guskov@ensam.eu} \\
\And
	\href{https://orcid.org/0000-0003-0469-8437}{\includegraphics[scale=0.06]{orcid.pdf}\hspace{1mm}}{Marc Rébillat}$^*$ \\
    Laboratoire PIMM\\
	Arts et Métiers Sciences et Technologies\\
	\texttt{marc.rebillat@ensam.eu} \\
\And
\href{https://orcid.org/0000-0002-3178-4587}{\includegraphics[scale=0.06]{orcid.pdf}\hspace{1mm}}{Nicolas Ranc}$^*$ \\
Laboratoire PIMM\\
Arts et Métiers Sciences et Technologies\\
\texttt{nicolas.ranc@ensam.eu}  \\
}

\renewcommand{\headeright}{Preprint}
\renewcommand{\shorttitle}{}

\hypersetup{
pdftitle={Exact identification of nonlinear dynamical systems by Trimmed Lasso},
pdfsubject={Machine Learning},
pdfauthor={Shawn L.~Kiser,  Mikhail Guskov, Nicolas Ranc, Marc Rébillat},
pdfkeywords={sparse regression, nonlinear dynamics, uncertainty quantification, model discovery, hysteresis},
}

\begin{document}
\maketitle

\begin{abstract}
	Identification of nonlinear dynamical systems has been popularized by sparse identification of the nonlinear dynamics (SINDy) via the sequentially thresholded least squares (STLS) algorithm. Many extensions SINDy have emerged in the literature to deal with experimental data which are finite in length and noisy. Recently, the computationally intensive method of ensembling bootstrapped SINDy models (E-SINDy) was proposed for model identification, handling finite, highly noisy data. While the extensions of SINDy are numerous, their sparsity-promoting estimators occasionally provide sparse approximations of the dynamics as opposed to exact recovery. Furthermore, these estimators suffer under multicollinearity, e.g.~the irrepresentable condition for the Lasso. In this paper, we demonstrate that the Trimmed Lasso for robust identification of models (TRIM) can provide exact recovery under more severe noise, finite data, and multicollinearity as opposed to E-SINDy. Additionally, the computational cost of TRIM is asymptotically equal to STLS since the sparsity parameter of the TRIM can be solved efficiently by convex solvers. We compare these methodologies on challenging nonlinear systems, specifically the Lorenz 63 system, the Bouc Wen oscillator from the nonlinear dynamics benchmark of No\"el and Schoukens, 2016, and a time delay system describing tool cutting dynamics. This study emphasizes the comparisons between STLS, reweighted $\ell_1$ minimization, and Trimmed Lasso in identification with respect to problems faced by practitioners: the problem of finite and noisy data, the performance of the sparse regression of when the library grows in dimension (multicollinearity), and automatic methods for choice of regularization parameters.
\end{abstract}
 

\section{Introduction}

Ideas from inverse modeling and data-driven system identification techniques are well situated for nonlinear dynamical systems in engineering domains. Classically, these techniques develop a nonparametric or parametric model of a physical system from coordinates, e.g.~measurements. Parametric models require full state and latent variables of the system, which can be unknown in experiments. To deal with this, nonparametric identification methods can be leveraged since they only require input-output measurements, e.g.~block-oriented Volterra  \citep{rebillatIdentificationCascadeHammerstein2011}, NARMAX \citep{billingsNonlinearSystemIdentification2013}, and neural network models \citep{masriApplicationNeuralNetworks2000}, since they are able to regress to functional mapping between inputs and outputs with high accuracies. However, they suffer from interpretability and the mappings can have no physical significance with respect to the dynamics of the underlying system. For that reason, identification of nonparametric models are often coined as a black-box method. 

An alternative semi-parametric method named the Sparse Identification of Nonlinear Dynamics (SINDy), is a data-driven model discovery framework introduced to identify the ordinary differential equations (ODEs) of dynamical systems in the form $ \dot{{z}}(t) = {f}({z}(t))$; with known measurable state variables ${z}(t) \in \mathbb{R}^M, \forall t \in [0, T]$, from candidate state-dependent functions ${f}({z}(t)) : \mathbb{R}^M\rightarrow \mathbb{R}^M$ \citep{bruntonDiscoveringGoverningEquations2016}. In discrete matrix vector-form the dynamics can be written for the $j$th state measurements of time-length $M$ as:
\begin{equation}
	\dot{\bm{z}}_j = \bm{\Theta}(\bm{z}) \bm{\xi}_j, \quad j = 1, \cdots, N \label{e1}
\end{equation} 
for a predetermined library matrix of $P$ candidate functions $\bm{\Theta} \in \{\mathbb{R}^{M\times P}:M\ge P\}$. This type of problem has equivalents in other domains, such as dictionary learning in signal processing or feature selection in machine learning. For ease of notation we drop the $(\bm{z})$ from the library matrix henceforth. The goal is to extract the true governing dynamics with $K$ sparsity:
\begin{equation}
	\hat{\bm{\xi}}_j = \argmin_{\bm{\xi}_j }  \bigl\{\  \|\bm{\Theta}\bm{\xi}_j - \dot{\bm{z}}_j \|_2^2 \  \bigr\} \quad \operatorname{s.t.} \quad \| \bm{\xi}_j\|_0 = K, \quad j = 1, \ldots, N \label{e2}
\end{equation}
where the $\ell_0$ pseudo-norm is the $\operatorname{card}(\bm{\xi}_j)$. However, \cref{e2}, known as the best subset problem \citep{millerSelectionSubsetsRegression1984}, is NP-hard and requires $\binom{P}{k} $ combinatorial search for unknown $k$. Thus, a computationally feasible form of the  $\ell_0$ pseudo-norm is highly sought for which selects the true sparse model, i.e.~$\mathbb{P}(\hat{\bm{\mathcal{S}}}_j = \bm{\mathcal{S}}^*_j  )\rightarrow 1 $, for an estimated support $\hat{\bm{\mathcal{S}}}_j \coloneqq \{1;\ \hat{\xi}_j[k] \neq 0,\, k = 1,\ldots,P\}$ and true support $\bm{\mathcal{S}}_j^* \coloneqq \{1;\ \xi_j^*[k] \neq 0,\, k = 1,\ldots,P\}$. This framework combines elements of both black-box and white-box modeling approaches, and is considered a grey-box method.

SINDy has been applied to a large variety of dynamical systems, including nonlinear \citep{lopesPhysicalConsistencyEvolution2022} and hysteretic \citep{laiSparseStructuralSystem2019} mechanical oscillators, the boundary volume problem of elastic beams \citep{sheaSINDyBVPSparseIdentification2021}, reduced-order models of fluid flows \citep{loiseauConstrainedSparseGalerkin2018a}, and multiple timescale dynamics \citep{bramburgerPoincareMapsMultiscale2020}. Application of SINDy leads to parsimonious and interpretable solutions but suffers from many practical aspects when applied experimentally:\begin{enumerate}
	\item SINDy requires the inclusion of potential functions to be provided to form the library's basis. Too large of a basis leads to a violation of correlation conditions, i.e.~multicollinearity for regression \citep{farrarMulticollinearityRegressionAnalysis1967} and sparse regression \citep{zhaoModelSelectionConsistency2006}. A large basis of many functions might be provided by practitioners in the case of completely unknown dynamics, which limits the applicability and reliability of sparse regression algorithms. Additionally, systems with unobservable latent variables, such as hysteresis, cannot be measured and their existence must be inferred.
	\item The effect of noise on SINDy occurs from two aspects: noise intrinsic to the measurement data and noise introduced when making numerical approximations of derivatives or integrals. For the former, a variety of methods such as local and global denoising techniques \citep{cortiellaPrioriDenoisingStrategies2022}, by using a more robust optimization procedure \citep{cortiellaSparseIdentificationNonlinear2021}, or by enforcing simultaneously denoising and system recovery in optimization \citep{https://doi.org/10.48550/arxiv.2203.13837} can be used. 
	
	For the latter, integral (weak) formulations of SINDy have been devised to mitigate noise propagated by numerical derivatives. The most obvious method \citep{schaefferSparseModelSelection2017} replaces the problem formulation by numerically estimating the integral formulation of SINDy. The method devised by Messenger and Bortz \citep{messengerWeakSINDyGalerkinBased2021} utilizes test functions for integration by parts to estimate the derivative, while the occupation kernel technique by Rosenfeld et al.~\citep{rosenfeldOccupationKernelsDensely2019} uses test functions for estimating the derivative via the fundamental theorem of calculus.

	\item The hyperparameters used in the SINDy framework  are intrinsically tied to the problem formulation and estimator. Thus, they are tuned by model selection algorithms: usually the minimization of an information criterion, through Bayesian or frequentist methods, or heuristically, see \citep{dingModelSelectionTechniques2018} for a good overview. However, these selection criteria are prone to interpretability: the work by \citep{manganModelSelectionDynamical2017} shown that minimization of information criteria is not always likely to select the exact model; the work by \citep{cortiellaSparseIdentificationNonlinear2021} shown that corner selection for L-curve techniques can fail when the L-curve has multiple corners. 
	
	\item In the SINDy framework, the initial values are neglected from identification: this may be appropriate when dynamic systems are weakly nonlinear, are not chaotic, and tend to a steady-state. In practice, the initial condition is not estimated and is provided, which equates to assuming that noise does not affect this initial condition.
	It is critical to acknowledge that  estimating the initial values have practical benefits when SINDy is used for model predictive control or forecasts.
\end{enumerate}

Problematic identification can arise from dynamics corrupted by noise, finite data lengths, and a highly correlated library matrix. Furthermore, the estimator and the model selection algorithm should allow for an interpretable way for tuning hyperparameters. Specifically, we are interested in the property of exactly recovering the subset of candidate functions which make up the dynamics of the nonlinear model, i.e.~the oracle variable selection $\mathbb{P}(\hat{\bm{\mathcal{S}}}_j = \bm{\mathcal{S}}^*_j  )\rightarrow 1 $.

The original SINDy algorithm employs the sequentially thresholded least squares (STLS) to solve \cref{e2}. However, the STLS method introduces bias against coefficients with low magnitudes due to an increasing threshold parameter. It can be likened to a backward selection greedy algorithm, where coefficients of candidate functions are hard thresholded during sequential least squares on smaller supports until convergence. If a true coefficient is prematurely thresholded during optimization, the corresponding governing equation cannot be recovered, e.g.~see \citep{boninsegnaSparseLearningStochastic2018}. To overcome these limitations, an extension of SINDy incorporates bootstrap ensembling techniques (E-SINDy). E-SINDy utilizes statistics from the bootstrapped models, leading to enhanced robustness in variable selection for finite and noisy data. In this study, we specifically refer to E-SINDy using the STLS estimator with a median bagging inclusion probability of $\approx 0.6$. E-SINDy was recently shown that its approximations converge to a Bayesian inference via Markov chain Monte Carlo solution with a horseshoe prior \citep{gaoConvergenceUncertaintyEstimates2023a}.

An alternative to the STLS stems from the Lasso formulation\footnote{It is worth noting that the Lasso formulation can incorporate physics-based constraints such as structural symmetry or energy-based constraints, as explored in \citep{loiseauConstrainedSparseGalerkin2018a}, but these must be known \textit{a priori} and are not focused on for this general study.}  which involves convex minimization of the squared $\ell_2$ norm of the residual along with a penalized $\ell_1$ norm. Literature in the statistical domain \citep{tibshiraniRegressionShrinkageSelection1996} and signal processing domain \citep{chenApplicationBasisPursuit1998} have characterized the theoretical performance of this convex penalty. One estimator with oracle properties is the Adaptive Lasso \citep{zouAdaptiveLassoIts2006} or iterative reweighted $\ell_1$ minimization (IRL1) \citep{candesStableSignalRecovery2006}, which uses a re-weighted $\ell_1$ penalty. The reweighting scheme adjusts the $\ell_1$ penalty by incorporating the individual coefficients from the previous iteration. The reweighted $\ell_1$ penalty is stage-wise convex allowing for efficient computation through homotopy continuation methods \citep{efronLeastAngleRegression2004,asifFastAccurateAlgorithms2013}. However, similar to STLS, increasing the penalty parameter in the Lasso variant leads to the shrinkage of smaller coefficients of candidate functions.

Greedy and convex $\ell_1$ penalty methods, including STLS and IRL1, occasionally achieve suboptimal sparse solutions since they approximate \cref{e2}. This includes results such as $\hat{\bm{\mathcal{S}}}_j \supseteq  \bm{\mathcal{S}}^*_j$, i.e.~a sparse solution that includes the true support but is over complete, and/or $\bm{\Theta}\bm{\xi}_j^* \approx \bm{\Theta}\hat{\bm{\xi}}_j: \|\hat{\bm{\xi}}_j \|_0 \neq K$, i.e.~a sparse solution that approximates the solution but does not satisfy the sparsity $K$.  Several other non-convex penalties have been proposed in the literature to close the gap between \cref{e2} and its approximations. These include separable penalties such as the smoothly clipped absolute deviation \citep{fanVariableSelectionNonconcave2001}, the Dantzig selector \citep{candesDantzigSelectorStatistical2007}, and the minimax concave penalty \citep{zhangNearlyUnbiasedVariable2010}. To maintain several desirable properties relevant to \cref{e2} and to promote further sparsity, non-separable non-convex penalties have been proposed such as sparse Bayesian learning \citep{wipfIterativeReweightedEll2010}, the moderately clipped Lasso \citep{kwonModeratelyClippedLASSO2015}, or the Mnet \citep{huangMnetMethodVariable2016}. 

In this study, we wish to introduce  the \underline{T}rimmed Lasso for \underline{r}obust \underline{i}dentification of \underline{m}odels (TRIM) for nonlinear dynamical initial value problems. 
Specifically, the Trimmed Lasso's non-convex penalty offers exact control over the desired level of sparsity for regression. Thus, its formulation is closer to \cref{e2} than previously mentioned convex methods for sparse estimation. While originally introduced over a decade ago in \citep{cohenCompressedSensingBest2009} and its name coined recently by \citep{bertsimasTrimmedLassoSparsity2017}, developments in optimization have made convex approximations of the Trimmed Lasso. TRIM has a very intuitive hyperparameter tuning process, which directly probes the Pareto front. While E-SINDy has recently been shown to have oracle properties in \citep{gaoConvergenceUncertaintyEstimates2023}, and the IRL1/Adaptive Lasso by \citep{zouAdaptiveLassoIts2006}, we will demonstrate across different dynamical systems that TRIM is postulated to have lenient conditions to satisfy its oracle properties, and thus better performance. 

Our contributions can be described as follows: We provide the methodology of TRIM for identification, and combined with the usage of wild bootstraps for uncertainty quantification for sparse nonlinear dynamic initial value problems under high noise, multicollinearity, and finite data length. We affirm more recent studies of sparse regression that Lasso-based formulations with non-convex penalties can achieve much better performance than the STLS used in SINDy and E-SINDy.  Additionally, uncertainty quantification comes after model section for TRIM, meaning that resampling techniques are performed for only one set of hyperparameters. This is a computational advantage over E-SINDy, which requires multiple bootstrapping models per hyperparameter, from which model selection criteria identifies the dynamics. Finally, we demonstrate TRIM on the chaotic Lorenz system, the Bouc Wen oscillator from the nonlinear dynamics benchmark of Noël and Schoukens, 2016, and identification and uncertainty quantification for a time delay system modeling tool cutting dynamics.

\paragraph{Notations} We denote \(\bm{x}\) and \(\bm{X}\) as vectors and matrices respectively. The $i$th column vector of a matrix is denoted \(\bm{X}_i\). A noise perturbed value \(x\) is denoted \(\tilde{x}\) while an estimate is denoted \(\hat{x} \).  The variance of the white Gaussian noise is denoted \(\sigma^2\). The \(\ell_p\) norm is denoted \(\|\bm{X}\|_p\). The Hadamard (element-wise) product and inner product between \(\bm{x},\bm{y}\) is denoted \(\bm{x}\circ \bm{y}\) and \( \langle\bm{x}, \bm{y} \rangle\) respectively.

\section{Preliminaries} \label{sec2}
In this section, we introduce preliminaries within the
 SINDy framework of nonlinear dynamical model identification. Specifically we mention two sparsity promoting regressions in the literature. Finally, data-preprocessing and sparsity promoting regressions require tuning of hyperparameters, which we elucidate via automatic methods. 

\subsection{SINDy}
SINDy is a data-driven model discovery framework which was originally introduced to identify the governing equations in the form of \cref{e2}. The underlying assumption is that the underlying dynamics $ {f}({z}(t))$ are linearizable by $P\ge 2$ sparse candidate functions, such that:
\begin{equation}
	\dot{{z}}(t) = {f}({z}(t)) \approx \sum_{i=1}^P \xi_i \theta_i({{z}}(t)) \label{e3} 
\end{equation}
where $\Theta = \{\theta_i(z(t)) \}$ is an over complete set of candidate functions. Thus, the goal is to determine the weights $\xi_i$ such that the balance between error and parsimony of the approximation is optimal, and ideally retrieving the true governing equations. The vector-matrix form of \cref{e2} can be expanded as:
\begin{equation*}
	\begin{gathered}
	\dot{\bm{z}}_j=\begin{bmatrix}
	\dot{x}_j\left(t_1\right), & \dot{x}_j\left(t_2\right), & \cdots, & \dot{x}_j\left(t_M\right)
\end{bmatrix}^\mathrm{T} \in \mathbb{R}^M; \\
	\bm{\Theta}(\bm{z})=\begin{bmatrix}
	\theta_1\left({z}(t_1)\right) & \theta_2\left({z}(t_1)\right) & \cdots & \theta_P\left({z}(t_1)\right) \\
	\theta_1\left({z}(t_2)\right) & \theta_2\left({z}(t_2)\right) & \cdots & \theta_P\left({z}(t_2)\right) \\
	\vdots & \vdots & \ddots & \vdots \\
	\theta_1\left({z}(t_M)\right) & \theta_2\left({z}(t_M)\right) & \cdots & \theta_P\left({z}(t_M)\right)
	\end{bmatrix} \in \mathbb{R}^{M \times P} ; \\
	\bm{\xi}_j=\begin{bmatrix}
	\xi_{j,1}, & \xi_{j,2}, & \cdots, & \xi_{j,P}
	\end{bmatrix}^\mathrm{T} \in \mathbb{R}^{P} 
	\end{gathered}
\end{equation*}
where the columns of the library matrix $\bm{\Theta}(\bm{z})$ are prescribed candidate functions (e.g.~$\bm{z}$, $\bm{z}^2$, $\sin(\bm{z})$). SINDy is deemed a semi-parametric method since its intended to accommodate a parametric candidate library with a non-parametric data-driven regression. In this study, we restrict ourselves to the case $M\ge P$, i.e.~a problem where the data lengths are larger than the potential candidate functions.  

\subsection{Sparsity promoting regression}

In order to approximate \cref{e2}, \citep{bruntonDiscoveringGoverningEquations2016} proposes the sequentially thresholded least squares (STLS) estimator. In summary, it applies a hard thresholding to the coefficients $|\bm{\Xi}|<\varphi$ and sequentially performs least squares on new subsets of the support:
\begin{equation}
	\begin{aligned}
		\mathcal{S}(\hat{\bm{\Xi}}^{(i)}) & =\big\{| \hat{\bm{\Xi}}^{(i)}| \geq \varphi\big\};\quad  \forall \varphi > 0,\\
		\hat{\bm{\Xi}}^{(i+1)}_\text{STLS} & =\argmin_{\bm{\Xi} \in \mathcal{S}^{(i)}} \bigl \{\ \|\bm{\Theta}\bm{\Xi} - \dot{\bm{Z}}\|_2^2 \ \bigr \} \quad \operatorname{s.t.} \quad \mathcal{S}^{(i)} = \mathcal{S}(\hat{\bm{\Xi}}^{(i)})
		\end{aligned} \label{e4}
\end{equation}
 for arbitrary $i$ iterations, where $\bm{\Xi} = \begin{bmatrix} \bm{\xi}_1 & \cdots & \bm{\xi}_P \end{bmatrix}$, $\dot{\bm{Z}} = \begin{bmatrix} \dot{\bm{z}}_1 & \cdots & \dot{\bm{z}}_N \end{bmatrix}$, and $\mathcal{S}^{(i)}$ is the $i$th support of $\hat{\bm{\Xi}}$ which satisfies the constraint.  However, STLS exhibits similarities to a backward selection greedy regressions where if a true coefficient is thresholded at an early stage during the iterative optimization, the true underlying equation cannot be recovered. Yet, STLS has been successfully applied to a large variety of dynamic systems, and only requires a single thresholding parameter $\zeta$. Additionally, \cref{e4} has a computational complexity of $\mathcal{O}(MP^2)$ per thresholding parameter, making it  efficient and easy to implement.

Alternatively, the Lasso estimator introduced by \citep{tibshiraniRegressionShrinkageSelection1996}, is well suited to sparse identification because of its property that it sets some coefficients exactly equal to
zero: 
\begin{equation}
	\hat{\bm{\xi}}_{\text{Lasso}, j} = \argmin_{\bm{\xi}_j }  \bigl\{\  \|\bm{\Theta}\bm{\xi}_j - \dot{\bm{z}}_j \|_2^2+ \lambda \| \bm{\xi}_j\|_1 \  \bigr\};\quad  \forall \lambda > 0\label{e5}
\end{equation}
where the $\ell_1$ term promotes the sparsity of the regression, the $\ell_2$ term keeps the solution close to the measurement data, and $\lambda$ is the regularization term which balances between the two. For the regularization path $0<\lambda_1 < \ldots < \lambda_\text{max}$ the solution path is piecewise linear, and thus solved efficiently by least angle regression \citep {efronLeastAngleRegression2004} or homotopy methods \citep{asifFastAccurateAlgorithms2013}. Thus, the Lasso is typically stated with a per iteration complexity of $\mathcal{O}(MP)$ per regularization parameter.

Recent works in statistical and compressed sensing domains has provided theoretical performance for the Lasso, assuming that $\bm{\xi}^*$ is sparse and abides by the "beta-min" conditions \citep{zouAdaptiveLassoIts2006}: to ensure that a sparse model is obtained where the true sparse model is in the subset ($\mathbb{P}(\hat{\bm{\mathcal{S}}}_j \supseteq \bm{\mathcal{S}}^*_j   )\rightarrow 1 $ as $(M\rightarrow \infty)$) the library matrix $\tilde{\bm{\Theta}}$ must abide by the restricted eigenvalue condition \citep{buhlmannStatisticsHighDimensionalData2011}; to ensure that the exact sparse model is obtained ($\mathbb{P}(\hat{\bm{\mathcal{S}}}_j = \bm{\mathcal{S}}^*_j  )\rightarrow 1 $ as $(M\rightarrow \infty)$), the library matrix must abide by the irrepresentable condition \citep{buhlmannStatisticsHighDimensionalData2011}. When these conditions are not met, the Lasso produces large bias for the non-zero coefficients as it continuously shrinks all coefficients toward zero.

To correct for this behavior, the Adaptive Lasso  \citep{zouAdaptiveLassoIts2006} or IRL1 \citep{candesStableSignalRecovery2006} replaces the $\ell_1$ with a penalized weight vector:
\begin{equation}
	\hat{\bm{\xi}}_{\text{IRL1},j}^{(i+1)} = \argmin_{\bm{\xi}_j }  \bigl\{\  \|\bm{\Theta}\bm{\xi}_j - \dot{\bm{z}}_j \|_2^2+ \lambda \|\hat{\bm{w}}_j^{(i)}\circ \bm{\xi}_j\|_1 \  \bigr\};\quad  \forall \lambda > 0\label{e6}
\end{equation}
where the weight vector is defined as a function of previous selection's coefficients:
\begin{equation}
	\hat{\bm{w}}_j^{(i)} \coloneqq \frac{1}{|\hat{\bm{\xi}}_{\text{IRL1},j}^{(i)}|^\gamma};\quad  \forall \gamma > 0\label{e7}
\end{equation}
for additional hyperparameters of $i$ iterations and $\gamma$. The IRL1 was recently featured  for nonlinear dynamical model identification by \citep{cortiellaSparseIdentificationNonlinear2021}. This extension of the Lasso has oracle variable selection under less strict properties than the Lasso. Note that if $|\hat{\bm{\xi}}_\text{Lasso}|$ is large, the Adaptive Lasso employs a smaller penalty (less shrinkage) which yields a smaller bias in the $\ell_2$ sense. Since the irrepresentable condition for the Lasso is hard to meet in practice, the Lasso's estimates in the first stage effectively reduces the number of false positives for the Adaptive Lasso in the subsequent stage. This is due to the fact that the Lasso has high probability of the screening property  \citep{buhlmannStatisticsHighDimensionalData2011} and that for $\hat{\bm{\xi}}_\text{Lasso} = 0 \rightarrow \hat{\bm{\xi}}_\text{ALasso} = 0$.

\subsection{Automatic hyperparameter tuning} \label{sec2.3}For all regularized problems in this study, a hyperparameter terms must be chosen in some optimal sense. These problems with a single hyperparameter $\lambda$ have the form:
\begin{equation}
	\argmin _{\bm{\beta}}\left\{\ \|\bm{X} \bm{\beta}-\bm{y}\|^2_2 + \mathcal{R}(\bm{\beta},\lambda)\ \right\} \label{e8}
\end{equation}
where $\mathcal{R}(\square)$ is an operator which applies a penalty on the solution for arbitrary $\bm{X}$, $\bm{\beta}$, $\bm{y}$. For model-based regressions, some proxy of the sparsity is weighed against the likelihood of the model. These parametric information criterion are general to the model and are well known, such as the Akaike information criterion (AIC) and its corrected variant (AICc), The Bayesian information criterion (BIC), to name a few. The AIC has been shown to be an \textit{efficient} selector, since it uses the Kullback-Leibler divergence to balance model complexity and goodness of fit.  The BIC instead tries to maximize the posterior model probability and thus is a \textit{consistent} selector. The general form for information criterion is evaluated for all possible models $\mathcal{M}$:
\begin{equation}
	\argmin_{ \hat{\bm{\beta}} \in \mathcal{M}}\left\{\ \|\bm{X} \hat{\bm{\beta}} -\bm{y}\|^2_2 + \Pi\sigma^2    \|\hat{\bm{\beta}}\|_0\ \right\} \label{e9}
\end{equation}
 where the left-hand side is the residual sum of squares and $\Pi$ is a stochastic penalty. Since the true $\sigma^2$ is unknown, the least squares estimate of ${\hat{\sigma}^2_\text{LS}} \coloneqq \|\bm{X} \hat{\bm{\beta}}_\text{LS} -\bm{y}\|^2_2 / (M-P)$ for normally distributed errors is commonly used instead. \cref{e9} can be shown to have the form of the AIC, BIC, and others with a linear penalty $\Pi$ in \cref{table1}. Since the intent is to identify the true models, i.e.~\textit{consistency}, we opt for usage of the corrected risk inflation criterion (RICc), which has shown superior performance in literature \citep{zhangModelSelectionProcedure2010,JMLR:v13:kim12a}. In short, the "risk" refers to the expected prediction error of a model; RIC adjusts the maximum increase in risk due to selecting predictors rather than knowing the "correct" predictors, and is proven to be a \textit{consistent} selector \citep{fosterRiskInflationCriterion1994}.

\begin{table}[ht]
	\centering \label{table1}
	\begin{tabular}{@{}ll@{}}
	\toprule
	Criterion & Stochastic penalty ($\Pi$) \\ \midrule
	AIC       &  $\Pi_\text{AIC}=2$  \\
	AICc      &  $\Pi_\text{AICc}=2 + \frac{2(\operatorname{card}( \bm{\beta} ) + 1)}{M-\operatorname{card}( \bm{\beta} )-1}$  \\
	BIC       &  $\Pi_\text{BIC}=\log M$  \\ 
	HQC       &  $\Pi_\text{HQC}=c \log\log M, \ \ \text{for } c>2$  \\
	RIC       &  $\Pi_\text{RIC}=2\log P$  \\
	RICc      &  $\Pi_\text{RICc}=2(\log P+\log \log P)$  \\	 \bottomrule
	\end{tabular}\vspace{0.25cm}
	\caption{Common penalty factors in parametric information criterions. Note the Hannan and Quinn criterion (HQC) requires a coefficient $c$ to be chosen, see \citep{hannanDeterminationOrderAutoregression1979}.}
\end{table}

When no model is known \textit{a priori}, e.g.~for denoising and derivatives estimations, nonparametric model selection techniques must be used. Commonly known are the leave-one-out and generalized cross-validation (GCV) methods. In this study, we propose an L-curve \citep{hansenAnalysisDiscreteIllPosed1992} be used for tuning $\lambda$. The L-curve represents a Pareto front on a log-log scale, with the penalty on the ordinate and the error on the abscissa:
\begin{equation}
	\mathcal{L}_\text{c}(\lambda)\coloneqq(a,\, o) \rightarrow\left\{\begin{array}{l}
		a\coloneqq\log \|\bm{X} \bm{\beta}-\bm{y}\|^2_2 \\
		o\coloneqq\log \mathcal{R}(\bm{\beta},\lambda)
		\end{array}\right. \label{e10}
\end{equation}
The L-curve has the name due to the fact an "L" shape is formed, occasionally with a well-defined corner point which would have a large curvature. Certain methods exist to find the maximum curvature, some notable being  \citep{hansenAnalysisDiscreteIllPosed1992,cultreraSimpleAlgorithmFind2020}. We opt for our own method which determines the maximum curvature through arc length approximations along the L-curve. 

While the L-curve is originally a heuristic tool in choosing the regularization parameter, a theoretical basis on why regularization parameters corner point works well has been studied in \citep{hansenAnalysisDiscreteIllPosed1992,reginskaRegularizationParameterDiscrete1996}. A recent Bayesian formulation of the regularization problem \citep{Antoni_2023}, demonstrates that the L-curve is a graphical way of searching for the maximum \textit{a posteriori} solution after marginalization over the priors. Additionally, the L-curve corner criterion has an advantage over the cross validation which presumes noise that is additive white Gaussian (AWGN) \citep{karlssonPerformancesModelSelection2019}. 

\begin{figure}[ht]\centering
	\includegraphics[width=\textwidth]{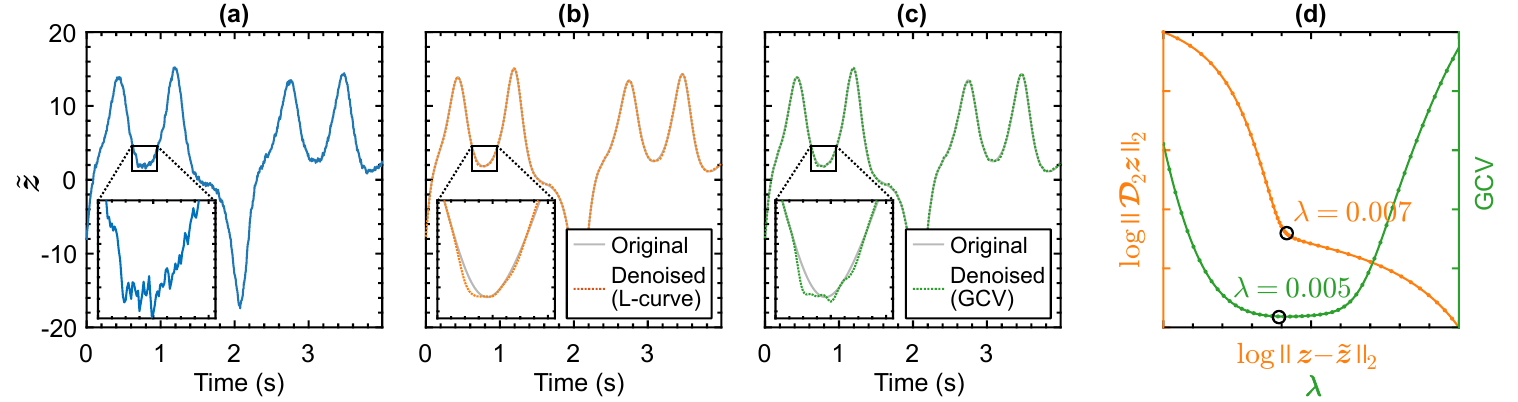}
	\vspace{-0.5cm}\caption{Demonstration of the effect of autocorrelation on model selection criterion for the nonparametric problem of the form of \cref{eb1}. In \textbf{(a)}, we simulate the popular Lorenz 63 system and corrupt the second degree of freedom (DOF) with correlated noise and plot its path. Tikhonov regularized denoising defined in \cref{eb1} is used to denoise the data, where the L-curve and GCV in \textbf{(d)} is used to determine the regularization parameter. Tikhonov denoising in \textbf{(b)} using the corner point of the L-curve (RMSE$ = 1.3 \cdot 10^{-3}$)  can denoise correlated noise better than the minimum of the GCV (RMSE$ = 1.9\cdot 10^{-3}$) in \textbf{(c)}. The full systematic parameters used for the simulation are given in \cref{appB}.}\label{figure1}
\end{figure}

To see this, consider the leave-one-out cross-validation set which has the expectation:
\begin{equation*}
	\mathbb{E} \big( (\bm{X} \hat{\bm{\beta}})^\text{out} - \tilde{\bm{y}} \big)^2=\mathbb{E}\big( ((\bm{X} \hat{\bm{\beta}})^\text{out} - \tilde{\bm{y}} )^2 \big)+\sigma^2 + 2 \operatorname{cov}\big((\bm{X} \hat{\bm{\beta}})^\text{out},\, \tilde{\bm{y}}\big) 
\end{equation*}
where $\mathbb{E}(\msquare)$ is the expectation operator, $\msquare^\text{out}$ is an estimate on a portion of data, and $\tilde{\bm{y}} = \bm{y}^* + \bm{\varepsilon}$ is the true data with AWGN of variance $\mathbb{E}(\bm{\varepsilon})= \sigma^2$. Cross-validation  explicitly assume that $\operatorname{cov}\big((\bm{X} \hat{\bm{\beta}})^\text{out},\, \tilde{\bm{y}}\big)\asymp \bm{0}$. Thus, if the regularized expression of \cref{e9} describes the denoising scenario, a smooth signal plus correlated noise will regress towards a solution of a less smooth signal plus white Gaussian noise. A denoising demonstration of this phenomena can be found in \cref{figure1}, where it's shown that the L-curve is the most robust when this assumption is violated.  In this sense, the L-curve has an advantage that it does not rely on an \textit{a priori} statistical measure; corner points are where the solution norm and penalty have a Pareto front. However, the L-curve fails when the discrete picard condition is not met \citep{hansenDiscretePicardCondition1990}, i.e.~the Fourier coefficients of the data $\tilde{\bm{y}}$ decay to zero faster than their singular values.

\section{TRIM overview}
In this section, we introduce the methodology of TRIM, first by introducing the non-convex penalty of the Trimmed Lasso, followed by an intuitive hyperparameter tuning method which directly probes the Pareto front and allows for a forward selection if higher accuracy is needed. Finally, TRIM extends upon the usual derivative form of SINDy to estimate the initial values of nonlinear dynamical models.

\subsection{The Trimmed Lasso}
For the problem of \cref{e2}, a solution that respects the $\ell_0$ pseudo-norm becomes computationally infeasible as $P$ grows. However, an attractive property of \cref{e2} is the ability to enforce the sparsity. An ideal penalty would be able to be penalized by $\lambda$ and enforce sparsity $K$ simultaneously, which would take the form of:
\begin{equation*}
		\hat{\bm{\xi}}_j = \argmin_{\bm{\xi}_j }  \bigl\{\  \|\bm{\Theta}\bm{\xi}_j - \dot{\bm{z}}_j \|_2^2 \ + \mathcal{R}(\bm{\xi}_{j},\lambda ) \bigr\} \quad \operatorname{s.t.} \quad \| \bm{\xi}_j\|_0 = K 
\end{equation*}
One such penalty, the Trimmed Lasso, has a non-convex penalty that satisfies both control of the sparsity and can be penalized. It is defined as \citep{cohenCompressedSensingBest2009}:
\begin{equation}
	\mathcal{T}(\bm{\xi}_j,k)=\sum_{i=k+1}^P\left|\xi_{j,i}\right| \label{e11}
\end{equation}
where $|\xi_{j,1} | \ge | \xi_{j,2}| \ge \ldots \ge |\xi_{j,P}| $ are the sorted descending entries of the coefficient vector. Intuitively, \cref{e11} penalizes the $k+1$ entries by the $\ell_1$ distance between $\bm{\xi}_j$ and the closest $k$ sparse vector. This forms a proxy for \cref{e2}:
\begin{equation}
	\hat{\bm{\xi}}_{\text{TRIM},j} = \argmin_{\bm{\xi}_j }  \bigl\{\  \|\bm{\Theta}\bm{\xi}_j - \dot{\bm{z}}_j \|_2^2 \ + \lambda \mathcal{T}(\bm{\xi}_j,k)\bigr\} ;\quad  \forall \lambda > 0\label{e12}
\end{equation} 
The Trimmed Lasso is a generalization of the standard Lasso, which recovers the Lasso when \(k=0\). For \(k>0\) and an increasing value of \(\lambda\), the shrinkage effect on the smallest \(P - k\) entries increases. When \(\lambda\) reaches or exceeds a threshold \(\acute{\lambda}\), the smallest \(P - k\) entries are forced to zero \cite{amirTrimmedLassoSparse2021}. This exactness property uniquely characterizes the Trimmed Lasso through the existence of \(\acute{\lambda}\), and for \(\lambda\) sufficiently large, \cref{e12} becomes a solution to the constrained best subset selection problem as defined in \cref{e2} \citep{amirTrimmedLassoSparse2021,gotohDCFormulationsAlgorithms2018}. 

However, if one sets \(\lambda\) arbitrarily large, so does the non-convexity of the problem. The optimization of \cref{e12} then is carried out by incrementally solving for \(\lambda\), where \(\lambda_1 < \lambda_2 < \ldots < \lambda_i < \ldots\). For each \(\lambda_i\) greater than or equal to \(\acute{\lambda}\), the estimate \(\hat{\bm{\xi}}_j(\lambda_i\ge \acute{\lambda})\) is projected onto the closest \(k\) sparse vector.  While the Trimmed Lasso has an intuitive but non-convex form, its ``convexification'' allows for efficient computation. An example of this can be formulated using an alternating minimization formulation \citep{bertsimasTrimmedLassoSparsity2017} of \cref{e12}:
\begin{equation}
	\begin{aligned}
	f(\bm{\xi}_j) & =\|\bm{\Theta}\bm{\xi}_j - \dot{\bm{z}}_j \|_2^2 / 2+ \lambda \mathcal{T}(\bm{\xi}_j,k) + \eta \|\bm{\xi}_j\|_1 \\
	f_1(\bm{\xi}_j) & =\|\bm{\Theta}\bm{\xi}_j - \dot{\bm{z}}_j \|_2^2 / 2+(\eta+\lambda)\|\bm{\xi}_j\|_1 \\
	\textstyle f_2(\bm{\xi}_j) &  \textstyle =  \langle \bm{\gamma},\,\bm{\xi}_j \rangle
	\end{aligned}\label{algeq1}
\end{equation}
such that $f(\bm{\xi}_j) \approx f_1(\bm{\xi}_j) - f_2(\bm{\xi}_j)$ and:
\begin{equation}
	\bm{\rchi}_j = \operatorname{arg\,sort}(|\bm{\xi}_j|);\quad\text{and}\quad\bm{\gamma}[{\rchi}_{j,i}] = \left\{\begin{aligned}
		\lambda \operatorname{sign}(\bm{\xi}_{j}[{\rchi}_{j,i}]) \quad &\text{if}\ i  \le k \\
		0 \quad &\text{if}\ i > k 
		\end{aligned}\right.; \quad \text{for}\ i = 1,\ldots,P  \label{algeq2}
\end{equation}
where $\operatorname{arg\,sort}(\cdot)$ returns a vector of descending index values and the square brackets indicate the vector entry with 1-indexing. Given the convexity of  $f_1(\bm{\xi}_j)$ and $f_2(\bm{\xi}_j)$,  the alternating minimization scheme of \cref{alg1} can be utilized. More efficient implementations have been algorithmically pursued via alternating direction method of multipliers (ADMM) \citep{bertsimasTrimmedLassoSparsity2017}, block coordinate descent \citep{yunTrimmingRegularizerStatistical2019}, and a surrogate penalty called the generalized soft-min coupled with the fast iterative shrinkage-thresholding algorithm (FISTA) \citep{amirTrimmedLassoSparse2021}. These implementations make \cref{e12} computationally feasible, with each bringing its own unique theoretical convergences and guarantees. 

\begin{algorithm}[H]
	\caption{Alternating minimization of Trimmed Lasso}
	\begin{algorithmic}[1] \label{alg1}
	\renewcommand{\algorithmicrequire}{\textbf{Input:}}
	\renewcommand{\algorithmicensure}{\textbf{Output:}}
	\Require $\ell_2$ norm scaled \(\breve{\bm{\Theta}}\) (see \cref{appC}), \(\dot{\bm{z}}\), sparsity \(k\), \(\nu\ge5\), \(\eta = 10^{-3}\), tolerance \(\varepsilon \)
    \Ensure  \(\breve{\bm{\xi}}_{\text{TRIM}}\)
	 \State $j \leftarrow 1$
	 \State \( \bm{\lambda} \leftarrow \exp(\verb|logspace|(\|\dot{\bm{z}}\|_2\cdot10^{-3},\,\|\dot{\bm{z}}\|_2 ,\, \nu)  )\) \Comment See \citep[
		Theorem 2.1]{amirTrimmedLassoSparse2021}.
		\For{\( i = 1 \) to \(\nu\)} 
			\State \(\lambda \leftarrow \bm{\lambda}_i ;\quad \breve{\bm{\xi}}^{(1)} \leftarrow \mathcal{N}(\bm{0},\bm{1}); \quad s \leftarrow 1 \)
			\While{ \( \|\bm{\Theta} \breve{\bm{\xi}}^{(s)} - \dot{\bm{z}} \|_2^2 / 2+ \lambda \mathcal{T}( \breve{\bm{\xi}}^{(s)},k) + \eta \| \breve{\bm{\xi}}^{(s)}\|_1 > \varepsilon \) }
			\State \(\bm{\gamma}^{(s)} \leftarrow \) \cref{algeq2} 
			\State  \(   \breve{\bm{\xi}}^{(s+1)} = \argmin_{\bm{\xi} }  \bigl\{\  \|\bm{\Theta}\bm{\xi} - \dot{\bm{z}} \|_2^2 / 2+(\eta+\lambda)\|\bm{\xi}\|_1  -  \langle \bm{\gamma}^{(s)},\,\bm{\xi} \rangle \bigr\} \) \Comment{Resolved using any Lasso solver.}
			\State \(s\leftarrow s+1\)
			\EndWhile
			\If{$\operatorname{card}(\breve{\bm{\xi}}^{(s)})=k$}
			\State \(\mathcal{M}(j) \leftarrow \breve{\bm{\xi}}^{(s)}  \) 
			\State  \( \mathcal{L}(j) \leftarrow \|\bm{\Theta} \breve{\bm{\xi}}^{(s)} - \dot{\bm{z}} \|_2^2 / 2+ \lambda \mathcal{T}( \breve{\bm{\xi}}^{(s)},k) + \eta \| \breve{\bm{\xi}}^{(s)}\|_1\)
			\State $j \leftarrow j +1$ 
			\EndIf
		\EndFor 
		\State	\(\breve{\bm{\xi}}_{\text{TRIM}}\leftarrow \mathcal{M}(i)\leftarrow \argmin_{i} \bigl\{  \mathcal{L}(i) \bigr\} \) \Comment{Take the support that gives the smallest residual.}
   \State \Return \(\breve{\bm{\xi}}_{\text{TRIM}}\) 
	\end{algorithmic} 
	\end{algorithm}

\begin{figure}[ht]\centering
	\includegraphics[width=\textwidth]{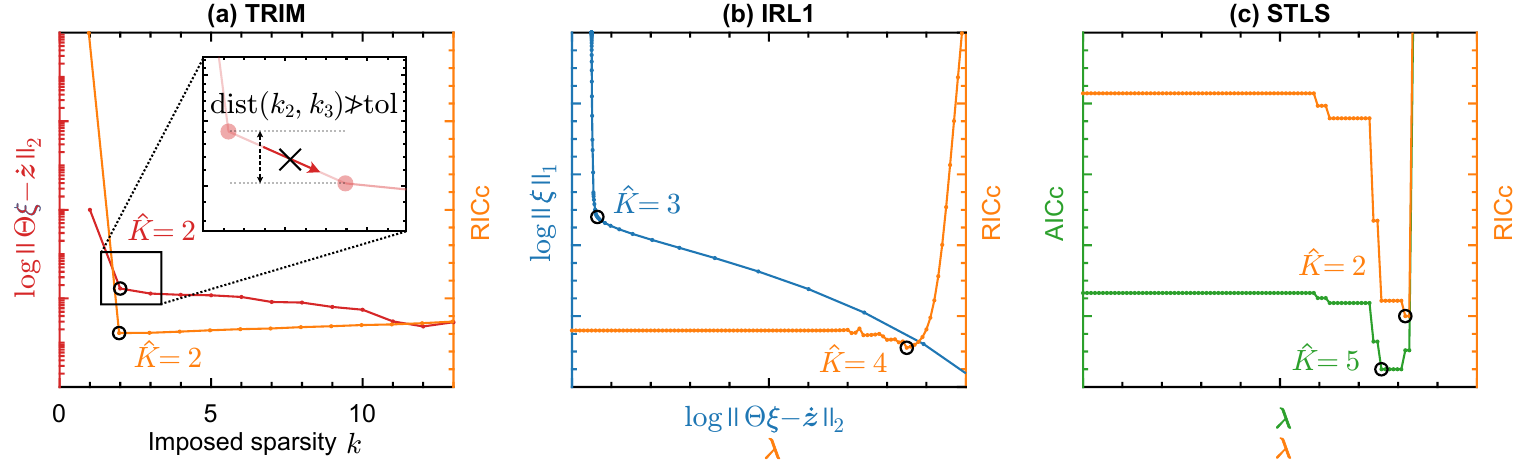}
	\vspace{-0.5cm}\caption{A demonstration of automatic model selection criteria paired with their respective sparse regression estimators for the noisy third DOF of the Lorenz 63 where the true sparsity is $K^*=2$. In \textbf{(a)}, TRIM's corner point in the L-curve defined by \cref{e13} as well as RICc achieves the true sparsity ($\hat{K}=2$). A forward step selection criteria is shown where an additional coefficient is permitted if the distance between the residuals satisfies a tolerance. In \textbf{(b)}, IRL1 obtains differing results: the L-curve method ($\hat{K}=3$) and the RICc ($\hat{K}=4$). In \textbf{(c)}, STLS obtains more consistent results with RICc ($\hat{K}=2$) and more efficient results with AICc ($\hat{K}=5$), confirming with theory in \cref{sec2.3}.  The full systematic parameters used for the simulation are given in \cref{appB}.
	}\label{figure2}
\end{figure}

This leads to the Trimmed Lasso having its sparsity $k$ as its single tunable hyperparameter. The easiest implementation is to simply utilize the information criterion approaches in \cref{e9}, namely RICc. Alternatively, we can utilize an augmented L-curve hyperparameter for tuning TRIM within the SINDy framework. This approach allows us to directly probe the Pareto frontier between the residual and the sparsity $k$:
\begin{equation}
	\mathcal{L}_{\text{TRIM},j}(k)\coloneqq(a,\, o) \rightarrow\left\{\begin{array}{l}
		a\coloneqq\log \|\bm{\Theta}\hat{\bm{\xi}}_j - \dot{\bm{z}}_j \|_2 \\
		o\coloneqq k
		\end{array}\right. \label{e13}
\end{equation}
If one does not want to rely on minimization of information criterion, \cref{e13} facilitates an intuitive process for hyperparameter tuning. As the desired level of sparsity $k$ increases, the solution residual generally decreases. The occurrence of a sudden drop in the residual is considered indicative of the parsimonious model, representing a corner point on the L-curve. Following the identification of this corner point, practitioners can further employ a forward step selection method, allowing for a trade-off between increased sparsity and a tolerated decrease in the residual distance between consecutive models (e.g., $\operatorname{dist}(k_i,k_{i+1})>\operatorname{tol}$). Conversely, the hyperparameter tuning procedure of the STLS and IRL1 estimators of \cref{sec2.3} increase the magnitude of their hyperparameter(s) to promote sparsity. It is important to note that their on-grid hyperparameters do not guarantee a $k$ sparse solution, whereas TRIM provides this guarantee. This automatic model selection process is demonstrated for the Lorenz system in \cref{figure2}.

\subsection{Initial value problem formulation}
Here we detail the initial value problem formulation for the SINDy framework. 
While the initial conditions for SINDy in the literature are often presented as known with exact certainty, this issue is frequently overlooked, whether intentionally or unintentionally. Specifically for TRIM, the classical derivative formulation of \cref{e1} is changed by adding the initial values $z_0$ to the problem formulation:
\begin{equation}
	\left\{\begin{array}{l}\displaystyle \dot{{z}}(t) = {f}({z}(t)) \approx \sum_{i=1}^P \xi_i \theta_i({{z}}(t)) \\
		{{z}}(t_0) = z_0\\[0.25cm]
	\end{array}\right.;\quad \forall t \ge t_0
		\label{e14} 
\end{equation}
In matrix-vector notation, the integral formulation of the problem can be formulated as:
\begin{equation}
	\bm{z}_j  = \bm{T}_1\bm{\Theta} \bm{\xi}_j + z_0\bm{1}_M + \bm{\varepsilon}_q 
\end{equation}
where $\bm{1}_M \in \mathbb{R}^M$ a vector of ones and $\bm{T}_1$ is a matrix integral operator defined in \cref{ea2}. The integral formulation of \citep{schaefferSparseModelSelection2017}, however, discards the initial values by regression on the augmented data $ \bm{z}_j'= \bm{z}_j - z_0\bm{1}_M$. Instead, we will treat the initial values as an additional coefficient. Using the Trimmed Lasso penalty and defining $\bm{\Gamma}\coloneqq \begin{bmatrix}\bm{1}_M & \bm{T}_1\bm{\Theta}\end{bmatrix}$, the corresponding minimization problem which minimizes the quadrature error is and allows initial value estimates is:
\begin{equation}
	\hat{\bm{\psi }}_{\text{TRIM}, j} = \argmin_{\bm{\psi}_j}  \bigl\{\  \|\bm{\Gamma} \bm{\psi}_j - {\bm{z}}_j \|_2^2+  \mathcal{T}(\bm{\psi}_j,k) \  \bigr\};\quad  \forall \lambda > 0 \label{e16}
\end{equation}
where $\hat{\bm{\psi }}_j = \begin{bmatrix} -\hat{z}_{0,j} & \hat{\bm{\xi}}_j	\end{bmatrix}^\mathrm{T}$ is the recovered initial values and the coefficients of the problem. The formulation for first-order ODEs benefits from estimating numerical quadrature instead  numerical derivatives.

\subsection{Uncertainty quantification}\label{sec3.3} To assess the sampling variability of an estimator, one can rely on bootstrap methods to generate confidence intervals on estimated coefficients. However, it's not known \textit{a priori} that measurement data are independent and identically distributed (i.i.d.), i.e.~are not autocorrelated, or when the estimates provided by model identification have residuals that are heteroscedastic, i.e.~not a constant variance in the residual of the model. Therefore, the wild bootstrap introduced by \citep{wuJackknifeBootstrapOther1986} builds upon the residual bootstrap by generating bootstrap samples that preserve any dependence structure and heteroscedasticity of the original data. These bootstrap samples are then used to estimate the sampling variability of an estimator and construct robust confidence intervals and their forecasts. 

Here, we detail the wild bootstrap process for a general estimator with an already identified model:

\begin{enumerate}
	\item Using the identified support of the model, compute the residuals $\hat{\bm{r}} = \bm{z}_j - \bm{\Theta} \hat{\bm{\xi }}_{j}$.
	\item Generate a bootstrap residual by resampling the residuals with replacement, denoted as $\hat{\bm{r}}'$.
	\item Construct a bootstrap data vector as $\bm{z}' =\bm{\Theta} \hat{\bm{\xi }}_{j} + \hat{\bm{r}}'$.
	\item Estimate the coefficients for the bootstrap sample with the imposed sparsity of $k$, denoted as $ \hat{\bm{\xi }}_{j}'$.
	\item Repeat steps 2-4 for many bootstrap iterations.
	\item Compute the bootstrap standard errors and construct confidence intervals based on the distribution of the bootstrap estimates.
\end{enumerate}

\section{Experimental study}\label{sec4}
In this section, we evaluate the performance of three sparsity-promoting estimators, paired with model selection criteria, in identifying nonlinear dynamical systems. Specifically, this includes the STLS estimator from the original SINDy\footnote{\url{https://faculty.washington.edu/sbrunton/sparsedynamics.zip}} \citep{bruntonDiscoveringGoverningEquations2016}, its ensembled counterpart E-SINDy\footnote{\url{https://github.com/urban-fasel/EnsembleSINDy}} \citep{faselEnsembleSINDyRobustSparse2022}, the IRL1 estimator\footnote{\url{https://intra.ece.ucr.edu/~sasif/homotopy/}} by \citep{asifFastAccurateAlgorithms2013}, and TRIM which implements the FISTA\footnote{\url{https://github.com/tal-amir/sparse-approximation-gsm}} implementation of the Trimmed Lasso. For IRL1, we fix the number of reweighting iterations to two, while keeping $\gamma$ and $\lambda$ tunable parameters. We will use SINDy and E-SINDy to denote the original and ensembled (median bagging) implementation of STLS. E-SINDy conducts 100 bootstrap resamples across all examples, requiring bootstrapping and ensembling for each thresholding parameter before performing model selection on the resultant ensembled models.

The default model selection criterion for each methods are: 
SINDy and E-SINDy with the model selection criterion RICc of \cref{e9}, IRL1  with the L-curve corner criterion of \cref{e10}, and TRIM with the L-curve criterion of \cref{e13}. Unless otherwise mentioned, the TRIM with the L-curve criterion includes a forward step selection tolerance of $\operatorname{tol} = 5\%$ with a distance measurement:
\begin{equation*}
	\operatorname{dist}(k_i,k_{i+1}) = \frac{\|\bm{\Theta}\hat{\bm{\xi}}_j(k_i) - \dot{\bm{z}}_j \|_2^2 }{\|\bm{\Theta}\hat{\bm{\xi}}_j(k_{i+1}) - \dot{\bm{z}}_j \|_2^2 }
\end{equation*}
These model selection criteria are chosen due to their optimal performance observed by the authors in this study as well as the theoretical justifications in \cref{sec2.3}. If not indicated differently, the hyperparameters for SINDy and E-SINDy are defined for a log-spaced grid of size 100, $\bm{\varphi}=\{10^{-3},\ldots, 10^{3}\}$, for IRL1 a log-spaced grid of size 75 $\bm{\lambda}=\{10^{-10},\ldots, 10^{6}\}$ and $\bm{q} = \{2, 2.5, \ldots 5 \}$, and for TRIM $\bm{k}=\{1,2,\ldots, 8\}$ and $\nu=10$.

All implementations of \cref{sec4} are programmed in MATLAB and provided on Github\footnote{\url{https://github.com/slkiser/TRIM}}. We apply TRIM to synthetic data of the Lorenz 63 system, the Bouc Wen oscillator from the nonlinear benchmark of \citep{noel2016hysteretic}, and a cutting tool dynamical time-delay system. Generally, it can be understood that successful identification heavily relies on the quality of the nonlinear system's state measurements. For all simulations, we recommend conditioning the library matrix to lower the condition number of the matrix, see \cref{appC}. To characterize the data and library matrix, we include the definition of noisy data:
\begin{equation*}
\tilde{\bm{z}} \coloneqq \bm{z}^* + \bm{\varepsilon}; \quad \bm{\varepsilon} \sim  \mathcal{N}(\bm{0},\sigma^2\bm{1}_M)
\end{equation*}
where the noise of variance $\mathbb{E}(\bm{\varepsilon})= \sigma^2$. To assemble the library matrix, we use the notation for an $n$ combinatorial polynomial, for exmaple:
\begin{equation}
	\mathcal{P}(\tilde{\bm{z}}_1,\tilde{\bm{z}}_2,n=2) = \begin{bmatrix} \tilde{\bm{z}}_1 & \tilde{\bm{z}}_2 & \tilde{\bm{z}}_1^2 & \tilde{\bm{z}}_2^2 &\tilde{\bm{z}}_1\circ \tilde{\bm{z}}_2
	\end{bmatrix} 
\end{equation}
to represent the generation of the library matrix through combinatorial candidates. For evaluation metrics, we include three measures: first the exact support recovery error:
\begin{equation}
	E_\mathcal{S} \coloneqq \mathbb{1}( \hat{\bm{\mathcal{S}}}  \cap \bm{\mathcal{S}}^* = K) 
\end{equation}
where $\mathbb{1}$ is the indicator function and $\bm{\mathcal{S}} = \begin{bmatrix} \bm{\mathcal{S}}_1 & \cdots & \bm{\mathcal{S}}_N \end{bmatrix}$ is the concatenated supports for each state measurement. As a reminder, different estimators obtain different rates of oracle recovery ($\mathbb{P}(\hat{\bm{\mathcal{S}}}_j = \bm{\mathcal{S}}^*_j  )\rightarrow 1 $ as $(M\rightarrow \infty)$) for conditions on the problem.  Second, we utilize a root mean square error (RMSE):
\begin{equation}
	\text{RMSE} \coloneqq 	\sqrt{ \frac{1}{MP} \sum_{j=1}^N \sum_{i=1}^P(\bm{\Theta}\hat{\bm{\Xi}} - \bm{\Theta}{\bm{\Xi}}^* )^2_{ji} }
\end{equation}
where  $\bm{\Xi} = \begin{bmatrix} \bm{\xi}_1 & \cdots & \bm{\xi}_P \end{bmatrix}$ is the column concatenation of all state measurements. The RMSE measures the difference in trajectory and measures the ability to forecast a non-chaotic dynamical system. Third, we use the coefficient error:
\begin{equation}
E_c \coloneqq 	\frac{\|\hat{\bm{\Xi}}-\bm{\Xi}^*\|_2}{\|\bm{\Xi}^*\|_2}
\end{equation}
We note that coefficient error is commonly used in the identification literatures but is a slightly misleading measure of accuracy, since slightly differing initial conditions and/or coefficient magnitudes can yield diverging RMSEs on trajectories.

\subsection{Lorenz 63 system}
The Lorenz 63 system is a well-known and widely studied model in the field of nonlinear chaotic dynamics. It was originally developed by Lorenz as a simplified representation of atmospheric convection. This system consists of a set of three ODEs that can form a distinctive butterfly-shaped attractor. Due to the strong nonlinear characteristics, predicting the future states of the Lorenz 63 system is challenging. The Lorenz 63 system to simulate is given as:
\begin{equation}
	\left\{\begin{array}{l}
	\dot{x}=\sigma(y-x)  \\
	\dot{y}=\rho x-y-x z  \\
	\dot{z}=x y-\beta z \\
	\end{array}\right.
	\end{equation}
where the coefficients that manifest chaos are given as $\sigma = 10$, $\rho = 28$, $\beta = 8/3=2.6\bar{66}$ and the initial conditions $\dot{x}(0) = -8$, $\dot{y}(0)=7$, $\dot{z}(0)=27$.  

In this example, we aim to demonstrate the robust support recovery capabilities of TRIM across a wide range of noise levels and finite data lengths. We conduct simulations of the Lorenz 63 system with Runge-Kutta via MATLAB's $\verb|ode45|$ with a time step of $\Delta t = 0.01$ for durations ranging from $\{1, 2 , \ldots, 10 \}$ seconds. To introduce noise, we corrupt each DOF's path and trajectory with additive white Gaussian noise (AWGN) at levels ranging from $\{0\%, 0.4\%, \ldots, 3.5\% \}$. The derivative formulation is used and the trajectory (derivative) is analytically provided and not estimated from the path data. Denoising is deliberately omitted to evaluate the estimators based solely on their noise robustness.

For the first and second tests, we use a library of combinatorial polynomial order $n=2,3$ respectively, such that ${\bm{\Theta}} = \begin{bmatrix} \mathcal{P}({\bm{x}},{\bm{y}},{\bm{z}},n) \end{bmatrix}$. This yields a library matrix with column size $P = 9,19$ respectively. Specifically, we run each scenario with 100 trials of AWGN and average the exact support recovery error $E_\mathcal{S}$ and the  coefficient error $E_c$. 

\begin{figure}[H]\centering
	\includegraphics[width=\textwidth]{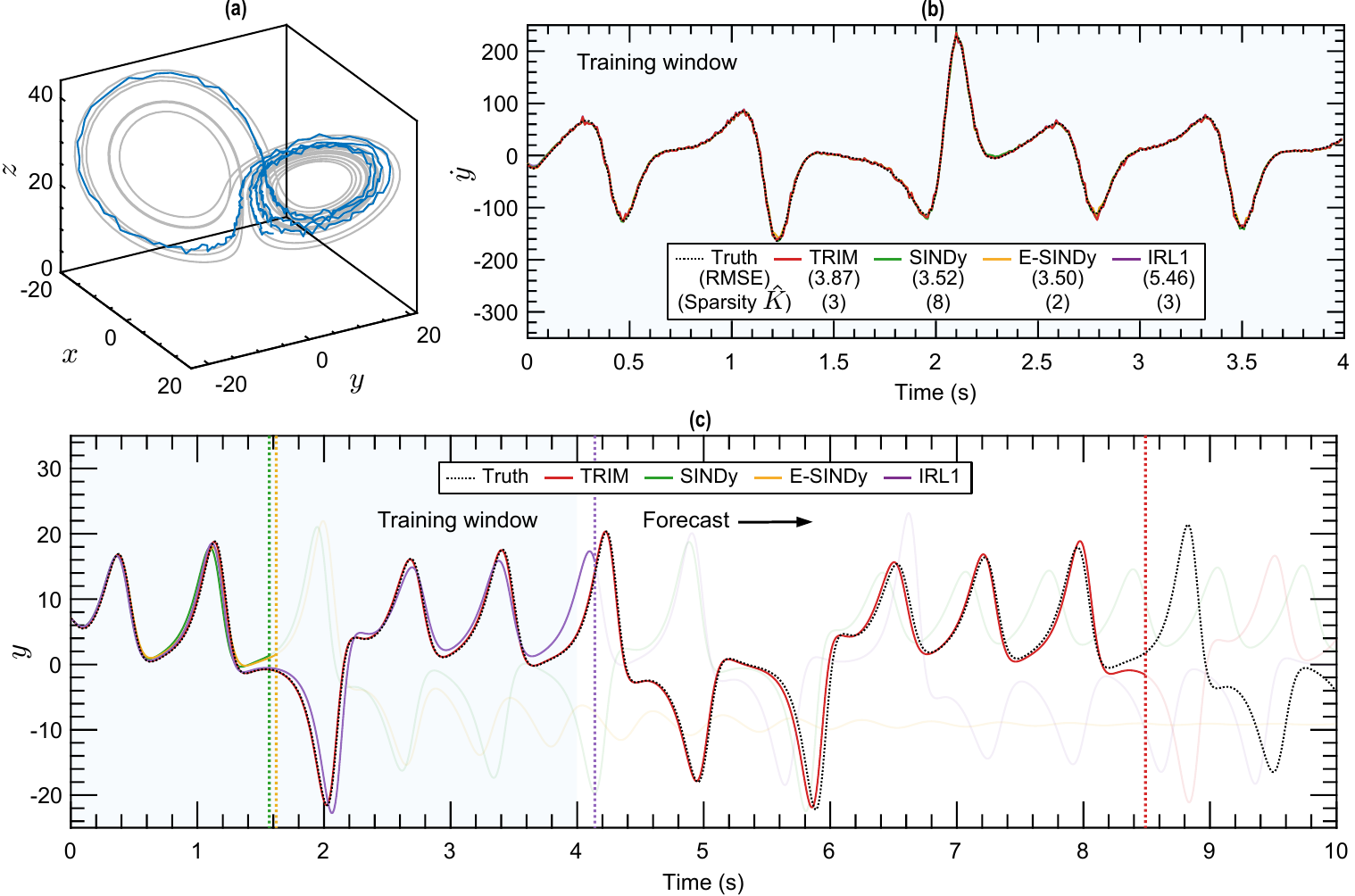}
	\vspace{-0.5cm}\caption{In \textbf{(a)}, the Lorenz 63 system is simulated for training data of $4$ seconds, $\Delta t = 0.01$, and corrupted with AWGN level of $2\%$. All estimators in \textbf{(b)} obtain sparse solutions that achieve low RMSE on the second DOF trajectory ($\dot{y}$) from a library of combinatorial polynomial of order $n=3$. However, TRIM  is capable of identifying exact model of the second DOF with sparsity $K^*=3$, whereas the other estimators fail.	This manifests when forecasting the second DOF path from the models identified: due to the chaotic dynamics, non-exact sparse models have highly divergent paths. The dashed vertical lines indicate when the residual of the forecast exceeds $10\%$ error.}\label{figure3}
\end{figure}

\begin{figure}[ht]\centering
		\includegraphics[width=\textwidth]{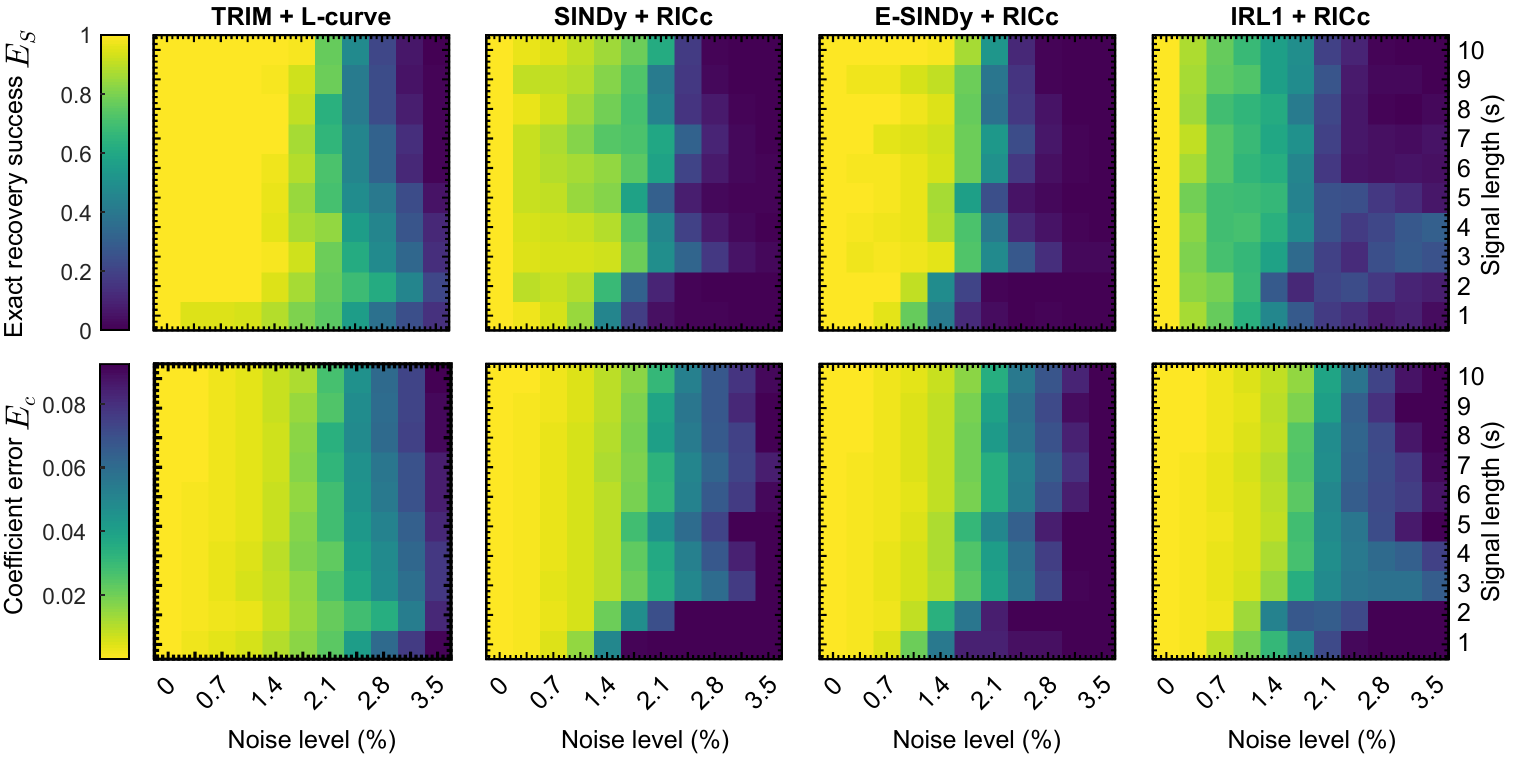}
		\vspace{-0.5cm}\caption{Sensitivity of TRIM, SINDy, E-SINDy, and IRL1 to noise level and data length in the Lorenz 63 system for a library matrix ${\bm{\Theta}} = \begin{bmatrix} \mathcal{P}({\bm{x}},{\bm{y}},{\bm{z}},n=2) \end{bmatrix}$. The time resolution is $\Delta t = 0.01$. The exact support recovery rate (top row) and coefficient error (bottom row) are averaged over 100 trials of AWGN per each noise and data length scenario.}\label{figure4}
	\end{figure}

Firstly in \cref{figure3}, we highlight a difficult scenario where other estimators fail compared to TRIM in forecast and exact sparse recovery. Using the derivative formulation, the estimators are given a library matrix of combinatorial polynomial order of $n=3$, training data of $4$ seconds, and AWGN at $2\%$.  TRIM is able to identify the correct model with exact sparsity $\hat{K}=3$ for the second DOF, whereas SINDy, E-SINDy, and IRL1 recover incorrect models with $\hat{K}=8,2,3$ respectively. As such, the path forecast using the true initial values of the Lorenz 63 system is reliably captured by the model identified by TRIM up to $\approx 8.5$ seconds, as opposed to only $\approx 1.6$ seconds by SINDy and E-SINDy and $\approx 4.2$ by IRL1. TRIM's robustness  is demonstrated by the heatmaps of exact support recovery error $E_\mathcal{S}$ and coefficient error $E_c$. For both the first and second tests the derivative formulation is employed. In \cref{figure4}, all estimators coupled with RICc perform relatively well for a combinatorial polynomial library $n=2$ with column size $P=9$. TRIM is shown to outperform in both metrics. E-SINDy enhances model recovery reliability and improves coefficient error compared to SINDy. However, both SINDy and E-SINDy share the same phase transition of failure at around $\approx 2.7\% - 2.8\%$ noise levels.

\begin{table}[ht]
	\centering
	\begin{tabular}{@{}ll@{}}
	\toprule
	Estimator & Identified model \\ \midrule
	TRIM      &       $	\left\{\begin{array}{l}
		\dot{x}=(10.044)\mathcolor{green}{y}-(10.047)\mathcolor{green}{x}  \\
		\dot{y}= (27.691)\mathcolor{green}{x}-(0.912)\mathcolor{green}{y}- (0.994)\mathcolor{green}{xz}  \\
		\dot{z}=(0.997)\mathcolor{green}{xy}- (2.667)\mathcolor{green}{z} \\
		\end{array}\right.$     \\ \hdashline[1pt/1pt]
	SINDy     &         $	\left\{\begin{array}{l}
		\dot{x}=(10.044)\mathcolor{green}{y}-(10.047)\mathcolor{green}{x}  \\
		\begin{aligned}\dot{y}= &(27.867)\mathcolor{green}{x}-(\mathcolor{red}{0})\mathcolor{green}{y}- (1.094)\mathcolor{green}{xz} -(0.108)yz\\& -(0.019)x^3 -(0.015)x^2y +(0.015)xy^2 + (0.004)xz^2 + (0.004)yz^2\end{aligned} \\
		\dot{z}=(0.997)\mathcolor{green}{xy}- (2.667)\mathcolor{green}{z} \\
		\end{array}\right.$         \\\hdashline[1pt/1pt]
	E-SINDy   &         $	\left\{\begin{array}{l}
		\dot{x}=(10.021)\mathcolor{green}{y}-(10.018)\mathcolor{green}{x}  \\
		\dot{y}= (25.482)\mathcolor{green}{x}-(\mathcolor{red}{0})\mathcolor{green}{y}- (0.950)\mathcolor{green}{xz}  \\
		\dot{z}=(0.998)\mathcolor{green}{xy}- (2.669)\mathcolor{green}{z} \\
		\end{array}\right.$        \\\hdashline[1pt/1pt]
	IRL1      &          $	\left\{\begin{array}{l}
		\dot{x}=(10.044)\mathcolor{green}{y}-(10.047)\mathcolor{green}{x}  \\
		\dot{y}= (26.451)\mathcolor{green}{x}-(\mathcolor{red}{0})\mathcolor{green}{y}- (0.956)\mathcolor{green}{xz} - (0.030){yz}  \\
		\dot{z}=(0.997)\mathcolor{green}{xy}- (2.667)\mathcolor{green}{z} \\
		\end{array}\right.$       \\ \bottomrule
	\end{tabular}\vspace{0.25cm}\caption{The identification results of TRIM, SINDy, E-SINDy, and IRL1 for the simulated Lorenz 63 system of \cref{figure3}. The correct functions are colored in green, missing coefficients are colored in red, and misidentified functions are colored in black. All estimators obtain the correct sparse solution for the first and third DOFs, but only TRIM obtains the correct second DOF with others obtaining a sparse approximation.
	}\label{table2}
	\end{table}

\begin{figure}[h]\centering
	\includegraphics[width=\textwidth]{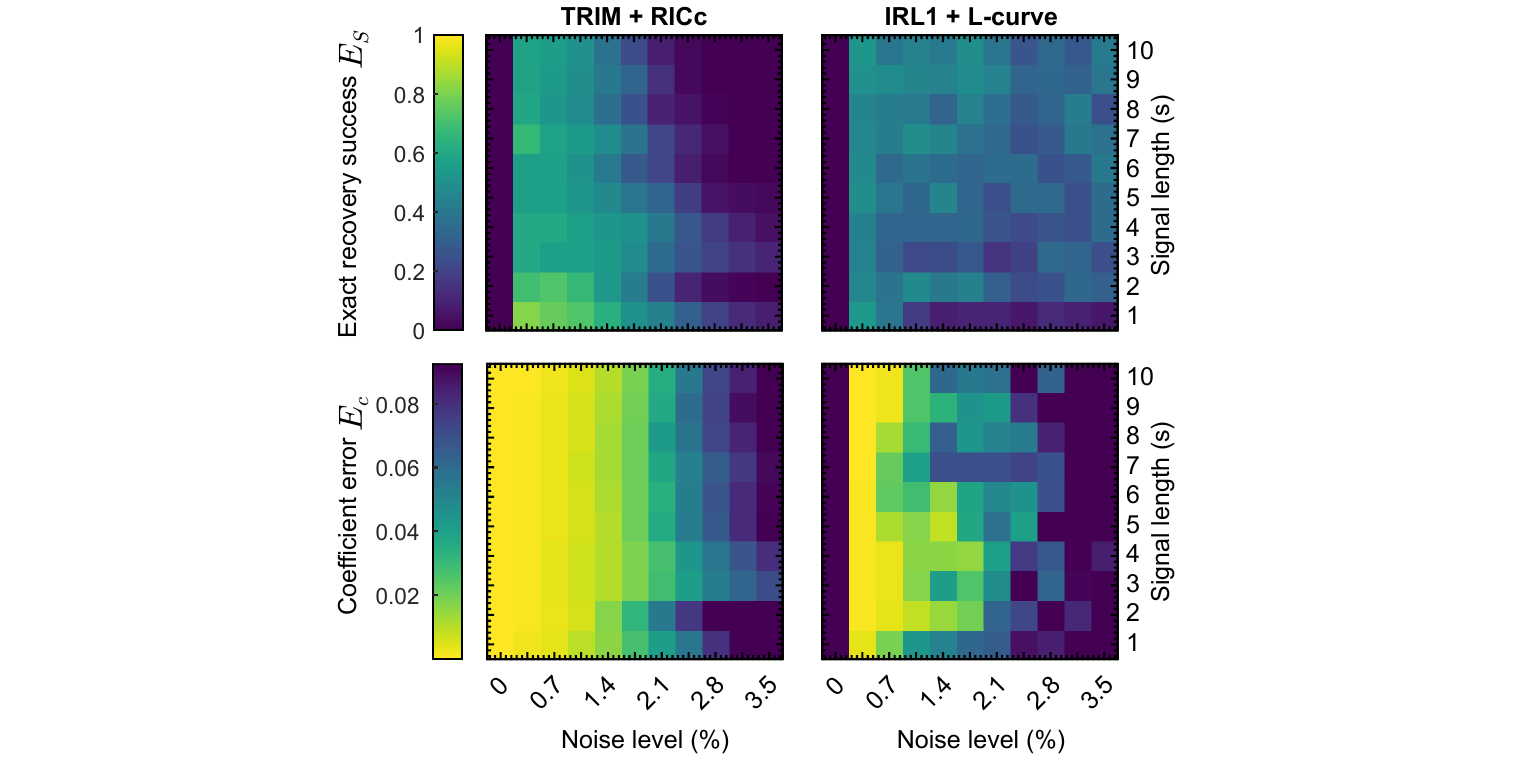}
	\vspace{-0.5cm}\caption{Sensitivity of TRIM and IRL1 using not recommended model selection criteria with respect to noise level and data length in the Lorenz 63 system for the same simulation parameters seen in \cref{figure4}. }\label{figure5}
\end{figure}

	In \cref{figure5}, we demonstrate that TRIM with RICc and IRL1 with L-curve criteria exhibit suboptimal exact support recovery rates. In the noiseless scenario, TRIM with RICc selects a sparse approximation due to lower residuals compared to the true model, while the absence of a corner point hampers the identification of the correct model with the L-curve criterion for IRL1. Henceforth, we avoid using these model selection criterion with their respective estimators.

\begin{figure}[H]\centering
	\includegraphics[width=\textwidth]{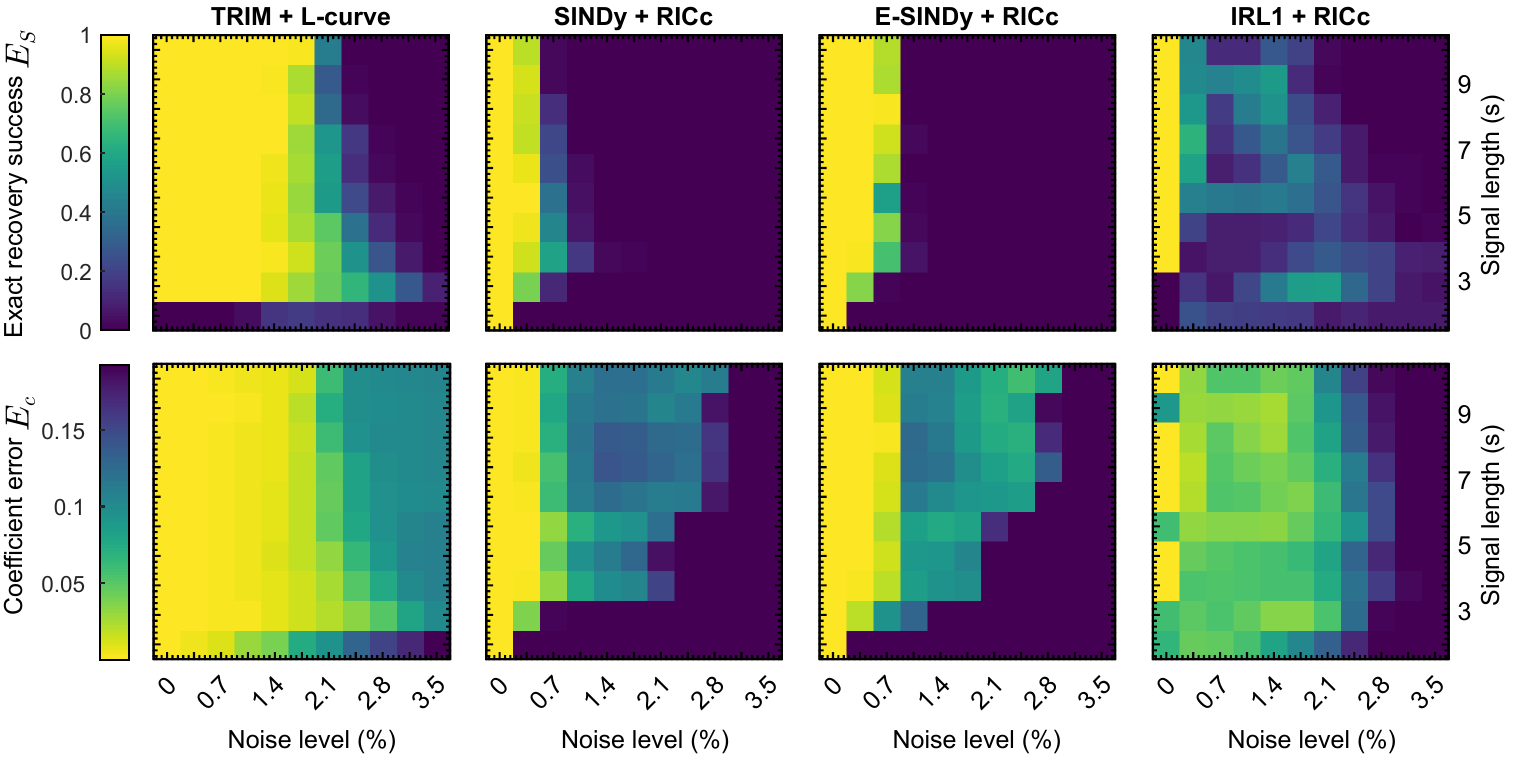}
	\vspace{-0.5cm}\caption{Sensitivity of TRIM, SINDy, E-SINDy, and IRL1 to noise level and data length in the Lorenz 63 system for a library matrix ${\bm{\Theta}} = \begin{bmatrix} \mathcal{P}({\bm{x}},{\bm{y}},{\bm{z}},n=3) \end{bmatrix}$. The time resolution is $\Delta t = 0.01$. The exact support recovery rate (top row) and coefficient error (bottom row) are averaged over 100 trials of AWGN per each noise and data length scenario.}\label{figure6}
\end{figure}

In \cref{figure6}, the effect of multicollinearity is amplified due to the increase in the dimension of the combinatorial polynomial library, where the library of order \(n=3\) has a column size of \(P=19\). The degeneration of all estimators can be seen, and the superiority of TRIM is emphasized in the phase transitions. Comparing \cref{figure4,figure6} conveys the effect of increasing the dimension of the library matrix, which is relevant for practitioners when the underlying functions are uncertain: TRIM is the most reliable at exact recovery when there is an increase of multicollinearity. 

\subsection{Bouc Wen hysteretic oscillator}\label{sec4.2}
The Bouc-Wen hysteresis oscillator is a widely studied model in the field of nonlinear dynamics and materials.  The behavior of the Bouc-Wen hysteresis nonlinearity is characterized by its ability to model complex, nonlinear phenomena such as rate-dependent effects, memory, and asymmetric responses. These phenomena manifest as hysteresis loops, which are a result of the system's input-output being dependent on both the current state and the history of the input. This system consists of a set of nonlinear differential equations:
\begin{equation}
	\left\{\begin{array}{l}
	m\ddot{x} + c \dot{x} + k{x} + z = u_\text{in} \\[0.1cm]
	\dot{z}=\alpha\dot{x}- \beta |\dot{x}||z|^{\nu-1}z-\delta \dot{x}|z|^{\nu}
	\end{array}\right. \label{e22}
	\end{equation}
where $m = 2$, $c=10$, $k = 5\cdot 10^4$ denote the linear mass, damping, and stiffness coefficients of a linear oscillator respectively, $u(t)$ the input, and $\alpha=5\cdot 10^4$, $\beta=8\cdot 10^2$, $\nu=1$, and $\delta=1.1\cdot 10^3$, which are the given parameters of the Bouc Wen nonlinearity to be identified from the benchmark \citep{noel2016hysteretic}. Additionally, the benchmark uses noiseless data and the initial conditions of the benchmark set $\dot{x}(0)=0$, $x(0) = 0$, and $z(0) = 0$. It is to be noted that the form given by \cref{e22} has redundant coefficients  \citep{maParameterAnalysisDifferential2004}, i.e.~the same behavior can be generated for multiple sets of coefficients. 

\begin{figure}[H]\centering
	\includegraphics[width=\textwidth]{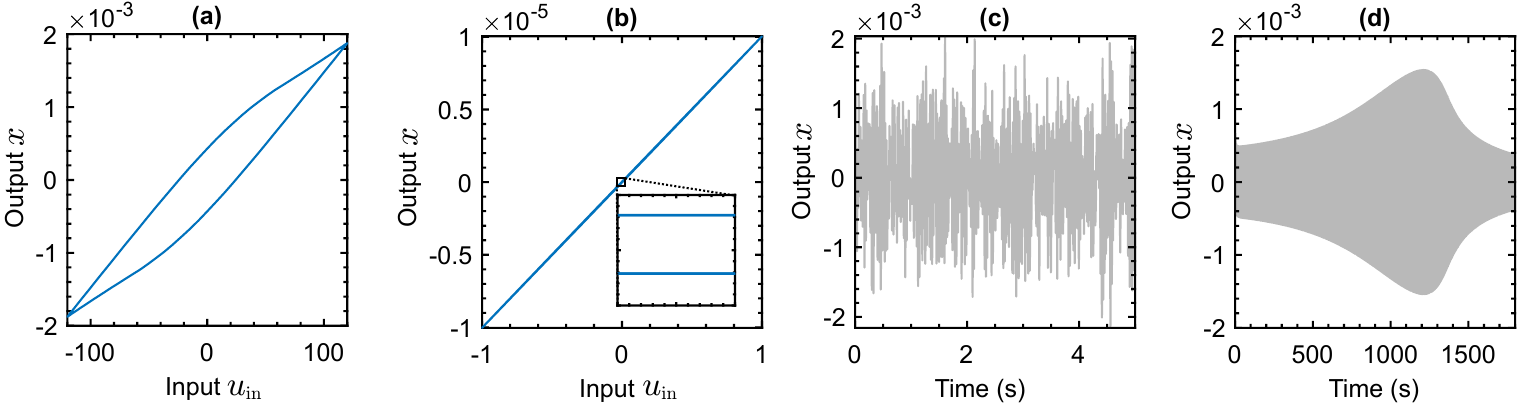}
	\vspace{-0.5cm}\caption{The Bouc Wen oscillator is simulated with a single sinusoidal input with amplitude $120$ Newton in \textbf{(a)} and $1$ Newton in \textbf{(b)}. The two datasets used for RMSE validation are provided by \citep{noel2016hysteretic}, where the outputs are shown: \textbf{(c)} is a multi-sine excitation and \textbf{(d)} a sine sweep. Their respective inputs are included by the benchmark but not shown. }\label{figure7}
\end{figure}

In this example, we aim to demonstrate that TRIM is well situated to recovering the exact support of hysteretic nonlinearities, especially of the Bouc Wen oscillator. This is validated using the two test datasets are presented in \citep{noel2016hysteretic} of purely input and output data\footnote{It should be reminded that the test datasets are not to be utilized during the process of training.} and identified models' RMSE is to be given as a figure of merit. We present two Bouc Wen input-output phase plot and the benchmark output data in \cref{figure7}.
This nonlinear system is challenging for identification since the internal variable $z$  is not a measurable quantity, and the absolute values to the power of $\nu$ do not allow Taylor or binomial expansions. Expanding on \citep{laiSparseStructuralSystem2019}, we show that hysteresis can easily be handled by TRIM with some algebraic manipulation. Consider \cref{e22} in which a mass-normalized linear oscillator is estimated \textit{a priori} to yield a discrepancy model:
\begin{equation}
	\left\{\begin{array}{l}
	f_\text{lin} + z =u_\text{in} \\[0.1cm]
	\dot{z}=\alpha\dot{x}- \beta |\dot{x}||z|^{\nu-1}z-\delta \dot{x}|z|^{\nu}
	\end{array}\right. \label{e23}
\end{equation}
where $f_\text{lin}$ is a linear approximation. To capture the latent history variable between input-outputs, we introduce the calculable state:
\begin{equation}
	y \coloneqq u_\text{in} - f_\text{lin} = z \label{e24}
\end{equation}
This yields the form to be identified:
\begin{equation}
	\dot{y}= \alpha\dot{x}- \beta |\dot{x}||y|^{\nu-1}y-\delta \dot{x}|y|^{\nu}
 \label{e25}
	\end{equation}
It can be seen that the memory and latent variables are captured by the difference between the external force and the linear oscillator. Note, to identify the second line of \cref{e25} via TRIM, the library matrix must have within the library of candidate functions $\{ \bm{y},\, \dot{\bm{x}} ,\, |\dot{\bm{x}}|\circ|\bm{y}|^{\nu-1} ,\, \dot{\bm{x}}\circ|\bm{y}|^\nu \}$. 

To address this Bouc Wen benchmark, we propose a three-step method\footnote{In the case that $f_\text{lin}$ must be estimated in one stage, i.e.~$m$, $c$, $k$ are to be identified in addition to the Bouc Wen coefficients simultaneously, then \cref{e25} can be expanded and solved in a more complicated fashion in \cref{appD}.} by separating the estimation of linear and nonlinear components. First, $\hat{f}_\text{lin}$ is estimated from the free decay of training data. Second, TRIM is used to identify the remaining Bouc Wen nonlinearity and initial conditions. Lastly, the linear oscillator's coefficients are estimated. The training data used is simulated via MATLAB's $\verb|ode45|$ with a single sinusoidal input with an amplitude of 50 Newtons, ending with a free decay, for a total length of 12 seconds, $\Delta t = 0.01$. We outline the steps taken as follows:
\begin{enumerate}
	\item Approximate $\hat{f}_\text{lin} = \hat{\ddot{x}} + 2\hat{\zeta}\hat{\omega}_n \hat{\dot{x}} + \hat{\omega}_n^2 x $, where the numerical derivatives are estimated using Tikhonov regularization defined in \cref{appC}, the damping ratio $\hat{\zeta}$ is estimated from the averaged log decrement of the free decay, and the damped natural frequency $\hat{\omega}_d$ from the maximum frequency peak using an appropriate super-resolution method \citep{kiserRealtimeSinusoidalParameter2023}.
	\item Use initial value problem formulation of TRIM (\cref{e16}) on $\dot{y}$ with the library $\bm{\Theta} = \begin{bmatrix} \mathcal{P}(\bm{x},|\bm{x}|,\dot{\bm{x}},|\dot{\bm{x}}|,\bm{y}, |\bm{y}|,n ) & \bm{u} \end{bmatrix}$, to identify the initial values and the coefficients and structure of the Bouc Wen model of $\hat{\dot{y}}$. For exact recovery, this yields $\{\hat{\dot{y}}(0), \hat{\alpha}, \hat{\beta}, \hat{\delta}, \hat{\nu} \}$.
	\item Estimate the linear oscillator's coefficients using a nonlinear regression, e.g.~a quasi-Newton method, on $m  \hat{\ddot{x}} + c \hat{\dot{x}} + k\hat{x} = u_\text{in}-\int \hat{\dot{y}}$, where the numerical integration of the estimated nonlinear model uses the discovered initial values $\hat{\dot{y}}(0)$. This yields $\{\hat{m}, \hat{c}, \hat{k}\}$.
\end{enumerate}
The role of the first step is to estimate the linear energy which is required for latent (memory) variables in a hysteretic nonlinearity.  For the second step, the library for sparse regression is assembled with candidate functions that one would expect when there is hysteretic behavior: $\bm{\Theta} = \begin{bmatrix} \mathcal{P}(\bm{x},|\bm{x}|,\dot{\bm{x}},|\dot{\bm{x}}|,\bm{y}, |\bm{y}|,n ) & \bm{u} \end{bmatrix}$. This is made with $n=2$ such that the library column space is $P=28$. The last step uses a nonlinear regression to refit the true coefficients of the problem. Note, that steps 1-3 can form an iterative method that refines linear and nonlinear coefficient estimates. 

We apply these steps to all three estimators and show their results of step 2 in \cref{table3}. When it comes to identifying the hysteretic nonlinearity in step 2, SINDy, E-SINDy, and IRL1 are unable to recover the exact model with sparsity $K^* = 6$ despite the training data being noiseless and providing a densely defined grid of hyperparameters. As a result, we omit the presentation of their results for the benchmark. We remark the that identified initial values corresponds to the error accumulated by the approximation of $\hat{f}_\text{lin}$. The initial conditions identified in step 3 do not align with the actual initial conditions; instead, they ensure the accuracy of the linear oscillator's coefficients. Finally, the estimated linear coefficients of TRIM's model are $\hat{m} = 2.037$, $\hat{c} =12.700$, and $\hat{k}=5.108\cdot 10^4$.

\begin{table}[h]
	\centering
	\begin{tabular}{@{}lll@{}}
	\toprule
	Estimator & Recovered model & Recovered I.C. \\ \midrule
	TRIM      &        $\dot{y}=(4.93\cdot 10^4) \mathcolor{green}{\dot{x}}- (8.16\cdot 10^2) \mathcolor{green}{|\dot{x}||y|^{\nu-1}y}- (1.16 \cdot 10^3) \mathcolor{green}{\dot{x}|y|^{\nu}}$    &   $\dot{y}(0)=-19.061$   \\ \cdashlinelr{1-3}
	SINDy     &         $\begin{aligned} \dot{y}=&(5.14\cdot 10^4) \mathcolor{green}{\dot{x}}- (1.04\cdot 10^3) \mathcolor{green}{|\dot{x}||y|^{\nu-1}y}- (1.17 \cdot 10^3) \mathcolor{green}{\dot{x}|y|^{\nu}} -(1.32\cdot 10^6)x \\ &  -(2.57\cdot 10^1)y	-(4.69\cdot 10^7)x|x| + (5.49\cdot 10^6)x|\dot{x}| -(1.85\cdot 10^3)x|y| \\ &   -(2.78\cdot 10^6)|x|\dot{x} + (1.91\cdot 10^3)|x|y - (3.32\cdot 10^5)\dot{x}|\dot{x}|	+ (1.57\cdot 10^{-2})y|y|	\\&+	(2.60\cdot 10^{1})u_\text{in}		\end{aligned}$   &   $\dot{y}(0)=-19.061$      \\\cdashlinelr{1-3}
	E-SINDy   &         $\begin{aligned} \dot{y}=&(4.89\cdot 10^4) \mathcolor{green}{\dot{x}}- (8.47\cdot 10^2) \mathcolor{green}{|\dot{x}||y|^{\nu-1}y}- (1.14 \cdot 10^3) \mathcolor{green}{\dot{x}|y|^{\nu}}-(7.21\cdot 10^5)\\ &  x -(1.39\cdot 10^1)y		-(1.23\cdot 10^3)x|y|	+ (8.68\cdot 10^2)|x|y	+(1.41\cdot 10^{1})u_\text{in}									\end{aligned}$        &   $\dot{y}(0)=-19.064$  \\\cdashlinelr{1-3}
	IRL1      &          $\begin{aligned} \dot{y}=&(4.92\cdot 10^4) \mathcolor{green}{\dot{x}}- (7.88\cdot 10^2) \mathcolor{green}{|\dot{x}||y|^{\nu-1}y}- (1.14 \cdot 10^3) \mathcolor{green}{\dot{x}|y|^{\nu}}+ (8.92\cdot 10^3) x\\ &  -(7.20\cdot 10^6)x|x|	-(2.08\cdot 10^6)x|\dot{x}|	 -(1.60\cdot 10^3)x|y|	+ (1.39\cdot 10^3)|x|y	\\&- (1.73\cdot 10^4)\dot{x}|\dot{x}|						\end{aligned}$   &   $\dot{y}(0)=-19.064$     \\ \bottomrule
	\end{tabular}\vspace{0.25cm}\caption{The identification results of TRIM, SINDy, E-SINDy, and IRL1 for the Bouc Wen hysteretic nonlinearity outlined in Step 2. The correct functions are colored in green and misidentified functions are colored in black.  While all estimators obtain a sparse solution that includes the true support $\hat{\bm{\mathcal{S}}}_j \supseteq  \bm{\mathcal{S}}^*_j$, only TRIM obtains the exact model with other estimators only obtaining a sparse approximation of the true Bouc Wen nonlinearity.
	}\label{table3}
	\end{table}

\begin{figure}[H]\centering
	\includegraphics[width=\textwidth]{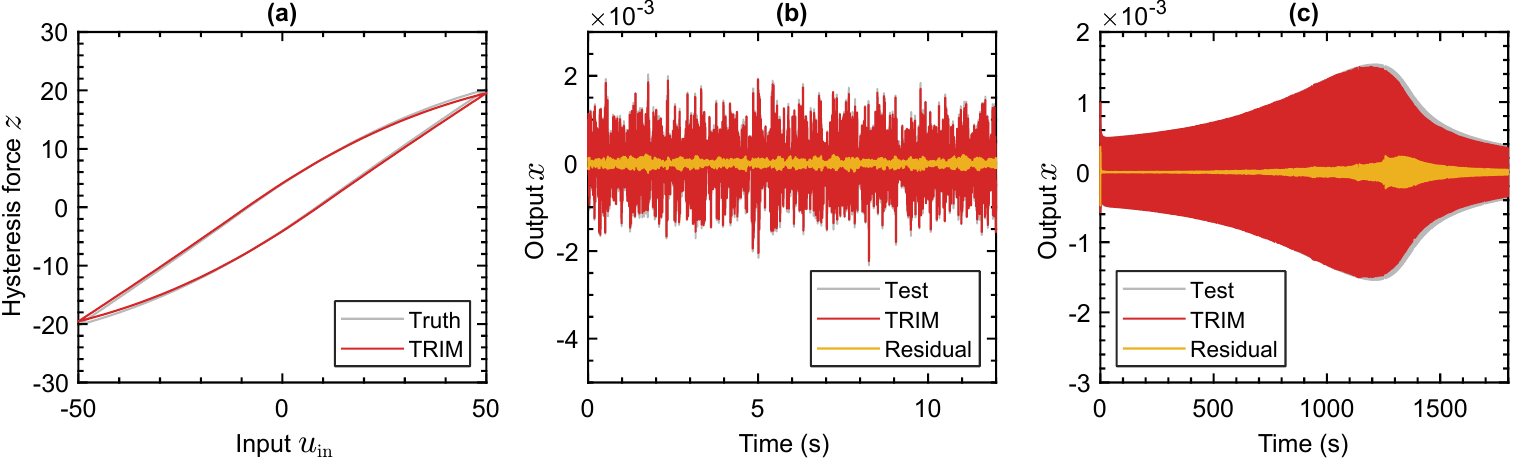}
	\vspace{-0.5cm}\caption{In \textbf{(a)}, the Bouc Wen training data is is simulated with a single sinusoidal input with amplitude $50$ Newton, where the true and TRIM estimated model's hysteretic force $z$ overlayed. The two datasets used for RMSE validation provided by \citep{noel2016hysteretic} are shown: \textbf{(c)} is a multi-sine excitation and \textbf{(d)} a sine sweep. The residuals between TRIM estimated model's and the test data output are shown in yellow. }\label{figure8}
\end{figure}

In \cref{figure8} we show the TRIM identified model's hysteretic force's phase plot on the training data and the simulation of two the two benchmark test datasets. Since TRIM recovers the exact support, the residual errors are simply due to coefficient and quadrature errors. Additionally, we show other benchmark results with respect to their RMSE and number of parameters in \cref{table4}; a majority of the other methods in the literature are black-box methods. TRIM can be shown to yield competitive results since it recovers the exact model, and its accuracy is only limited by its method of numerical integration. This is shown by the result notated ``Oracle'', where the true results are integrated using the provided MATLAB p-file from the benchmark \citep{noel2016hysteretic} with the provided test external forcing data. 
TRIM has the advantage of accurately identifying the precise governing equation and coefficients while being applicable to systems with general nonlinearities. However, its effectiveness is dependent on the availability of data capturing the free-decay behavior of the system.

\begin{table}[H]
	\centering
	\begin{tabular}{@{}lllll@{}}
	\toprule
	Estimator                                                                                   & \begin{tabular}[c]{@{}l@{}}RMSE Multi-sine\\ ($\cdot 10^{-5}$)\end{tabular} & \begin{tabular}[c]{@{}l@{}}RMSE Sine Sweep\\ ($\cdot 10^{-5}$)\end{tabular} & \begin{tabular}[c]{@{}l@{}}Parameters\\ (Sparsity $\hat{K}$)\end{tabular} & Exact recovery \\ \midrule
	\textbf{TRIM}                                                                                        & \textbf{6.569   }                                                              & \textbf{4.949}                                                                 & \textbf{6}          & \cmark             \\
	Volterra feedback \citep{schoukens2016modeling}                     & 8.409                                                                 & 5.601                                                                 & 14         & \xmark              \\
	Decoupled NARX \citep{westwickUsingDecouplingMethods2018}                  & 5.360                                                                  & 1.670                                                                  & 206        &  \xmark            \\
	EHH NN \citep{xuEfficientHingingHyperplanes2020}                           & 4.949                                                                & 2.402                                                                & 436        &  \xmark             \\
    LSTM \citep{schusslerDeepRecurrentNeural2019}                        & 5.980                                                                  & 2.800                                                                  & 21730      &  \xmark              \\
	MIMO PNLSS \citep{fakhrizadehesfahaniParameterReductionNonlinear2018}      & 1.871                                                                & 1.202                                                              & 217        &  \xmark            \\
	Decoupled PNLSS \citep{fakhrizadehesfahaniParameterReductionNonlinear2018} & 1.338                                                                & 1.117                                                                & 51         &  \xmark             \\
 \cdashlinelr{1-5}
	Oracle                                                                                      & 5.098                                                                 & 4.182                                                                 & 6          & -              \\ \bottomrule
	\end{tabular}\vspace{0.25cm}\caption{Results from TRIM and the literature for the Bouc Wen benchmark of \citep{noel2016hysteretic}. The "Oracle" entry refers to the RMSE found when using the true Bouc Wen parameters and the provided input data for simulation using the MATLAB p-file provided by the benchmark. Some acronyms are given: NARX refers to NARMAX without the moving average models; EHH NN stands for efficient hinging hyperplanes neural network, which is a wide but single hidden layer neural network; LSTM stands for a specific deep recurrent neural network (RNN) which uses  long short-term memory layers; MIMO stands for multiple-in-multiple-out; PNLSS stands for  polynomial nonlinear state space model.
	}\label{table4}
	\end{table}

\subsection{Self-excited vibration  (regenerative chatter) in machining system}
\begin{figure}[h]\centering
	\includegraphics[width=\textwidth]{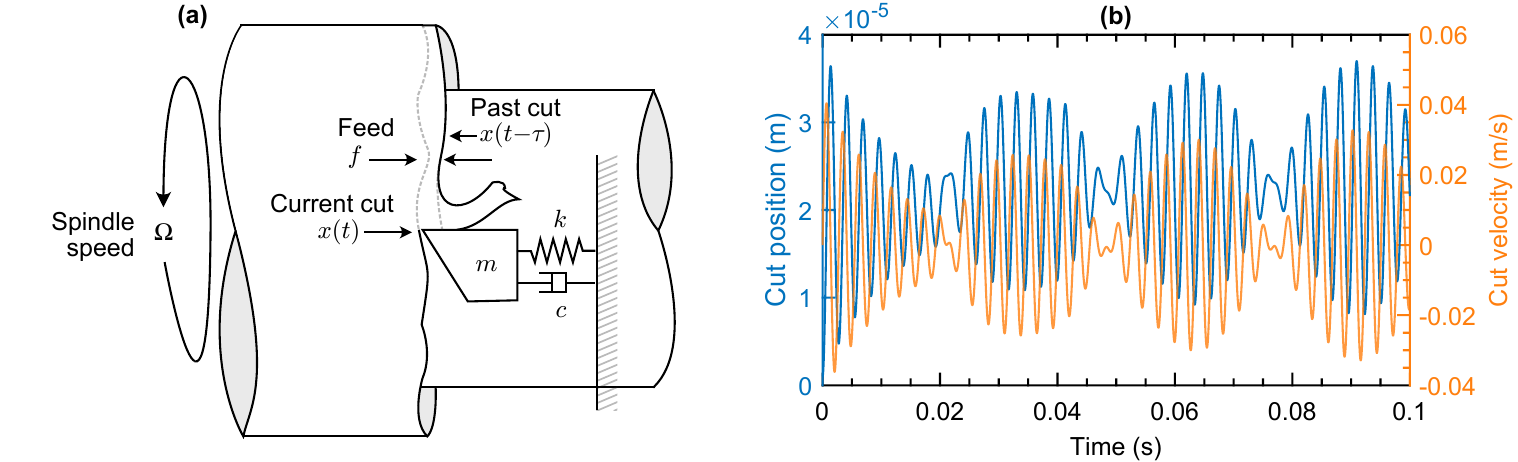}
	\vspace{-0.5cm}\caption{In \textbf{(a)}, the schematic oscillator model of self-excited chatter during a turning operation. In \textbf{(b)}, the simulated cut position and velocity for the unstable coefficients given in Table 5. }\label{figure9}
\end{figure}

The dynamics of metal cutting processes are often studied in the context to improve machining. One such predicament which occurs during metal cutting comes from unstable and erratic tool vibrations, namely regenerative chatter. This lead to poor surface quality, reduced tool life, and potentially damaging effects on the workpiece and cutting tool. A simple model of a turning process can be modeled as a mass-spring oscillator model with time delay. This is schematized in \cref{figure9}, where the dynamic interactions between the cutting tool and workpiece are parametrized. The delay differential equation (DDE) with constant delay  is given by \citep{altintasStabilityRegenerativeChatter2012}:
\begin{equation}
		\frac{1}{\omega_\text{n}^2}\ddot{x}(t) + \frac{2\zeta}{\omega_\text{n}} \dot{x}(t)+ 
		 x(t) = 
		\kappa(f  - x(t) + x(t-\tau)) - \rho \dot{x}(t)
 \label{e26}
\end{equation}
where  $\kappa$ is a nondimensionalized depth of cut coefficient in the feed direction, $f$ is the feed per revolution, and $\rho \coloneqq   \frac{\kappa C_\text{i}}{V K_\text{f}}$ is a coefficient that incorporates process damping $C_\text{i}$, cutting speed $V$, and cutting force coefficients from \citep{altintasStabilityRegenerativeChatter2012}. The state measurements correspond to the current cut surface, $x(t)$ and the previously cut surface  $ x(t - \tau )$. Typically, the feed  $f$ and time delay $\tau \coloneqq 1/\Omega$ are known since they are set by the machinist. 

In this example, our aim is to depict the scenario of estimating the stability lobes from a single acquisition of cutting data. To do so, the form of \cref{e26} must be identified, given the measured position data $x(t)$, where the feed and time delay are known. We extract confidence intervals from synthetic experimental data to inform the uncertainty bounds on the stability lobes, i.e., the critical stability of the machining process against free parameters. To admit \cref{e26} into state-space form, the acceleration is isolated such that:
\begin{equation}
	\ddot{x}  = (\kappa f  \omega_\text{n}^2)[1] - (\omega_\text{n}^2 +\kappa\omega_\text{n}^2) x -(2\zeta\omega_\text{n}+\rho\, \omega_\text{n}^2) \dot{x} +
	(\kappa \omega_\text{n}^2) x_\tau 
\label{e27}
\end{equation}

Experimental data is simulated using MATLAB's $\verb|dde23|$ with the initial history on the interval given as $x(t) = 10^{-6}: -\tau \le t \le 0$, $x(0) = 0$, a time discretization of $\Delta t = 10^{-5}$, and the coefficients given in Table 5. The velocity and acceleration is found using numerical differentiation, as suggested in \cref{appC}. Finally, the data is then corrupted with AWGN and denoised, resulting in an SNR of $\approx40$ dB. The library matrix is generated using combinatorial polynomial $n=3$ with $\bm{\Theta} = \begin{bmatrix} \mathcal{P}(\bm{x}(t),\dot{\bm{x}}(t),n ) & \mathcal{P}(\bm{x}(t-\tau),n) \end{bmatrix}$ yielding a library of column size $P=13$. Models are identified via TRIM and E-SINDy using the derivative formulation, with uncertainty bounds for TRIM established through the wild bootstrap's confidence intervals outlined in \cref{sec3.3}.

\begin{table}[h]
	\centering \label{table5}
	\begin{tabular}{@{}rrll@{}}
	\toprule
							  & \textbf{Parameter}   & \textbf{Coefficient} & \textbf{Value} \\ \midrule
	{Known}            & Time delay      &         $\tau$           & $2.000\cdot10^{-2}$              \\
							  & Feed rate       &      $f$              & $3.000\cdot10^{-3}$              \\ \cdashlinelr{1-4}
	{To be identified} 
							  & Damping ratio         &       $\zeta$             & $2.380 \cdot{10^{-2}} $          \\
							  & Natural frequency      &       $\omega_\text{n}$             & $2.129\cdot{10^3}$              \\
							  & Depth of cut   &        $\kappa$            & $1.500\cdot{10^{-1}}$           \\
							  & Process damping &        $\rho$            & $1.247 \cdot 10^{-5}$            \\\bottomrule
	\end{tabular} \vspace{0.25cm}  \caption{
		Coefficient values used to simulate the unstable self-excited chatter system, taken from \citep[page~139]{altintasStabilityRegenerativeChatter2012}.}
	\end{table}

\begin{figure}[H]\centering
	\includegraphics[width=\textwidth]{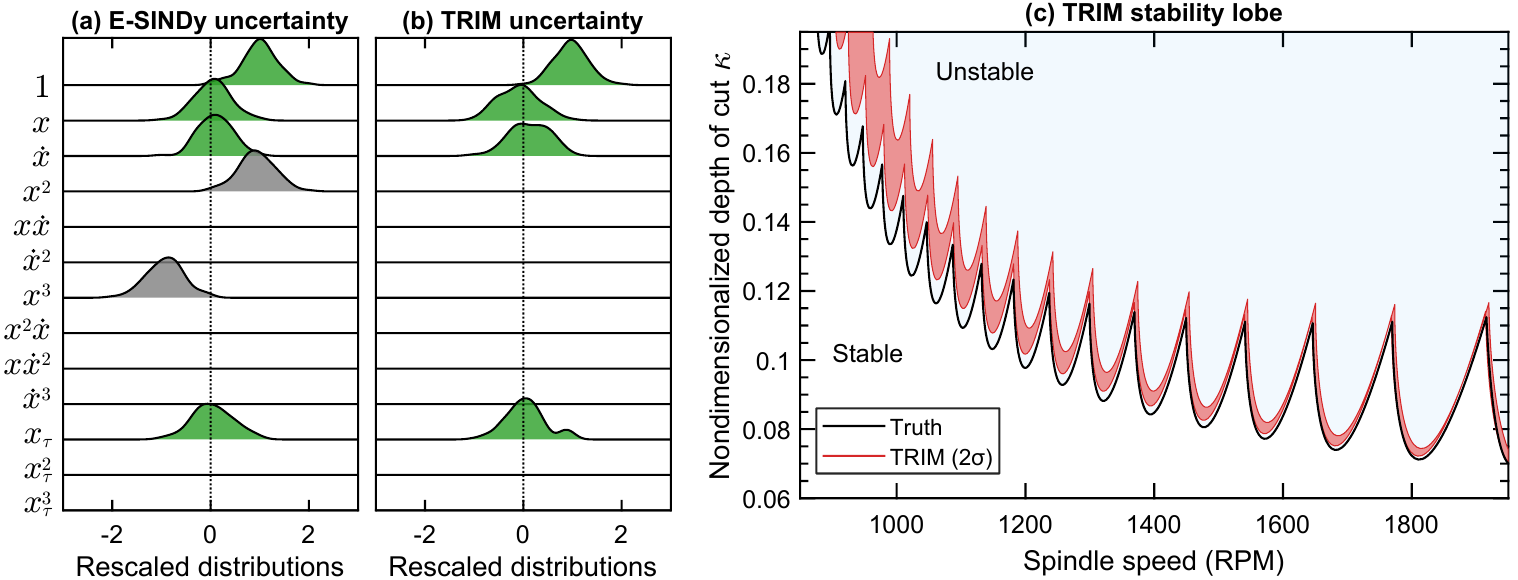}
	\vspace{-0.5cm}\caption{In \textbf{(a)} and \textbf{(b)}, E-SINDy and TRIM coefficients' rescaled cumulative distributions are shown as a ridgeline plot respectively, with 0 indicating the true value. Correctly identified coefficients are colored in green, whereas misidentified coefficients are colored in grey. In \textbf{(c)}, the true stability lobe diagram is juxtaposed with TRIM's stability lobe identified from one hundred  wild bootstraps, whose bounds are propagated coefficient uncertainties (5th and 95th percentiles, $2\sigma$). }\label{figure10}
\end{figure}

Upon performing uncertainty quantification on both TRIM and E-SINDy, one hundred bootstrapped iterations are used to derive statistics for the coefficients. Each coefficient's cumulative distribution is fitted using MATLAB's $\verb|fitdist|$, and the median model and uncertainties are visualized in \cref{table6} and \cref{figure10} respectively.  The coefficients are rescaled from $[-1,1]$ for clarity and differ in two scenarios:  from $[0,2]$ when they are $5\%$ larger than the true coefficient; and from $[-2,0]$ when they are $5\%$ less than the true coefficient. TRIM's coefficients demonstrate a concentrated distribution, accurately capturing the true model with $K^*=4$ terms of \cref{e27}. In contrast, E-SINDy's coefficients similar standard deviations but yielding an incorrect and non-sparse model, regardless of the information criteria utilized in \cref{sec2.3}. Consequently, these  results give a non-physical stability lobe diagram and is omitted from this paper.

\begin{table}[ht]
	\centering
	\begin{tabular}{@{}ll@{}}
	\toprule
	Estimator & Identified model with the  coefficients' median \\ \midrule
	TRIM      &       $	\ddot{x}  = (1.020\cdot 10^{2})[\mathcolor{green}{1}] - (5.211\cdot 10^{6} ) \mathcolor{green}{x} -(1.617\cdot 10^{2}) \mathcolor{green}{\dot{x}} +
	(6.765\cdot 10^{5}) \mathcolor{green}{x_\tau}$     \\ \hdashline[1pt/1pt]
	E-SINDy   &         $ \begin{aligned}	\ddot{x}  \,= &(1.005 \cdot 10^{2})[\mathcolor{green}{1}] - (4.982\cdot 10^{6}) \mathcolor{green}{x} - (1.619\cdot 10^{2}) \mathcolor{green}{\dot{x}} +
	(6.782\cdot 10^{5}) \mathcolor{green}{x_\tau} \\
	& - (1.098\cdot 10^{10})x^2 + (6.782\cdot 10^{14})x^3
	\end{aligned}$        \\ \bottomrule
	\end{tabular}\vspace{0.25cm}\caption{The median (50th percentile) of one hundred bootstrapped identification results of TRIM and E-SINDy for the simulated self-excited chatter system of \cref{figure9}. The correct functions are colored in green and misidentified functions are colored in black. TRIM obtains the correct sparse model whereas E-SINDy obtains a sparse approximation of the true model.
	}\label{table6}
	\end{table}

In order to propagate coefficient uncertainties onto the stability lobe diagram, we adapt the state-space formulation of \citep{chenComputationalStabilityAnalysis1997} for free parameters of spindle speed $\Omega = 1/\tau$ and nondimensionalized cutting depth $\kappa$:
\begin{equation}
	\begin{bmatrix}
		\dot{{x}}(t)\\\ddot{{x}}(t)
	\end{bmatrix} = \underbrace{\begin{bmatrix}
		0 & 1 \\ -(\omega_\text{n}^2 +\kappa\omega_\text{n}^2)\frac{1}{\Omega^2} & -(2\zeta\omega_\text{n}+\rho\, \omega_\text{n}^2)\frac{1}{\Omega}
	\end{bmatrix}}_{\bm{A}_1}\begin{bmatrix}
		{{x}}(t)\\\dot{{x}}(t)
	\end{bmatrix} + \underbrace{\begin{bmatrix}
		0 & 0 \\ \kappa \omega_\text{n}^2\frac{1}{\Omega^2} & 0
	\end{bmatrix}}_{\bm{A}_2}\begin{bmatrix}
		{{x}}(t-\tau)\\\dot{{x}}(t-\tau)
	\end{bmatrix}
		\label{e28}
\end{equation}
where the stability is found by roots of the characteristic equation:
\begin{equation}
\operatorname{det}(\mathrm{j}\omega\bm{I}- \bm{A}_1 - e^{-j2\pi\mu}\bm{A}_2) \quad \operatorname{s.t.} \quad \frac{1}{2} \le \mu \le 1 \label{e29}
\end{equation}
where $\omega$ is the chatter frequency and $\mu$ is related to the fractional vibration cycle in one complete rotation of the workpiece. The remaining details on the computational solution of \cref{e29} are relegated to \citep{chenComputationalStabilityAnalysis1997}. The 5th and 95th percentiles of the discovered coefficients are utilized in \cref{e29} where both the lower bound and upper bounds ($2\sigma$) of TRIM's stability lobes are superimposed on \cref{figure10}. At lower spindle speeds, it can be seen that TRIM overestimates the process damping coefficient, which manifests  at low spindle speeds with the stability lobes shifted rightwards. At elevated spindle speeds, TRIM converges towards the true stability lobes. This suggests that TRIM serves as a viable approach for identification of regenerative chatter systems, and more generally constant time-delay DDEs.

\section{Conclusions}\label{sec5}

SINDy is an effective framework that unites expert knowledge and data-driven sparse regression for the identification of nonlinear dynamical systems. The estimators presented in the original SINDy, namely STLS, or recently suggested alternatives like IRL1, offer computationally attractive ways to identify a sparse model of the dynamics. However, these estimators may fail to discern the true model under finite, noisy data, or high multicollinearity in the library matrix. E-SINDy, an extension to SINDy, was designed for robustness against finite and noisy data via bootstrapping and ensembling, has been demonstrated to have limitations. Its effectiveness hinges on the premise that variance reduction through ensembling outweighs bias introduced by bootstrapped models resampled from data -- a balance not consistently achieved when the model is misidentified.

TRIM, which is the non-convex penalty of the Trimmed Lasso paired with a simple Pareto-based L-curve model selection criteria, obtains exact recovery ($\hat{\bm{\mathcal{S}}}  \cap \bm{\mathcal{S}}^* = K^*$) for a wide range of dynamical systems. Furthermore, the penalty employs an intuitive constraint of only identifying $k$ terms, that more closely mirrors the $\ell_0$ pseudo-norm than the $\ell_1$ norm and its counterparts. TRIM's efficiency is further boosted by the ``convexification'' of the Trimmed Lasso, rendering it a fast, competitive solution in comparison to other sparsity-promoting estimators.
  
TRIM is shown to attain exact recovery under more robust conditions than other estimators, and still offers sparse approximation, like the IRL1 and STLS estimators, when these conditions are not met. We further illustrate that a variety of nonlinear dynamical systems can be adapted for sparse identification. Despite TRIM's successful identifications in the Lorenz 63 system, Bouc Wen hysteretic benchmark, and regenerative chatter system, the exploration of additional model selection criteria with respect to Trimmed Lasso is warranted. This rationale stems from TRIM's tendency to obtain monotonically decreasing residuals versus sparsity, and high likelihood of the true model residing near the L-corner. This suggests future studies into suitable criteria for adversarial scenarios lacking an L-shaped curve.

\section*{Acknowledgments}
We thank the authors Alexandre Cortiella and Zhilu Lai for providing their respective codes for reproduction. We also thank the Cassiopée high performance computing team (Boris Piotrowski, Théophile Gross, Camille Burton, and Paolo Errante) for enabling computations used in the study that were otherwise infeasible on personal workstations.

\appendix
\counterwithin*{equation}{section}
\renewcommand\theequation{\thesection\arabic{equation}}

\section{Integration and differentiation matrix operators} \label{appA}
The matrix form of the forward first and $i+1$th discrete differential operators, $\bm{D}_1\in \mathbb{R}^{M-1\times M}$ and $\bm{D}_{i+1}\in \mathbb{R}^{M-i\times M}$ respectively, are given by:
\begin{equation}
	\bm{D}_1 \coloneqq \frac{1}{\Delta t}\begin{bmatrix}
			-1 & 1 &  &  &    \\
			 & -1 & 1 &  &    \\
			 &  & \ddots & \ddots &    \\
			 &  &  &  -1 & 1 
	\end{bmatrix}; \quad\bm{D}_{i+1} \coloneqq\bm{D}_1 \bm{D}_i \label{ea1}
\end{equation}
where $\Delta t =  T/(M-1): t \in [0,T]$.  For numerical integrals, we use a matrix form of the first and $i+1$th cumulative integral operators, $\bm{T}_1,\,\bm{T}_i \in \mathbb{R}^{M \times M}$:
\begin{equation}
	\bm{T}_1 \coloneqq \bm{L}\bm{B}^{(\text{order})}; \quad \bm{T}_{i+1} \coloneqq \bm{T}_1 \bm{T}_i\label{ea2}
\end{equation}
where $\bm{L}\in\mathbb{R}^{M \times M}$ is a zero-padded lower triangular matrix and $\bm{B}^\text{(order)}\in\mathbb{R}^{M \times M}$ are the Newton polynomial matrices, e.g.~$\bm{B}^{(1)}$ corresponding to trapezoidal rule. 
The lower triangular matrix is given as:
\begin{equation*}
	\bm{L} = \begin{bmatrix}
		0&0 & 0 & 0 & \cdots & 0 & 0 \\
		0&1 & 0 & 0 & \cdots & 0 & 0 \\
		0&1 & 1 & 0 & \cdots & 0 & 0 \\
		0&1 & 1 & 1 &\cdots& 0 & 0 \\
		\vdots &\vdots & \vdots & \vdots& \ddots &\vdots &\vdots  \\
		0&1 & 1 & 1 & \cdots & 1 & 0 \\
		0&1 & 1 & 1 & \cdots & 1 & 1
		\end{bmatrix} 
\end{equation*}
The first-order Newton polynomial with forward difference gives the classic trapezoidal rule, whereas the third-order Newton polynomial gives Simpson's rule. We give the explicit matrices for the first three orders of polynomials:
\begin{equation*}
	\bm{B}^{(1)} =  \frac{\Delta t}{2}\begin{bmatrix}
		0 & 0 & 0& &\\
		1 & 1 & 0& & \\
			0 & 1 & 1 & & \\
			& \ddots & \ddots & \ddots & \\
			& &  0& 1 & 1
		\end{bmatrix}; \quad \bm{B}^{(2)} =  \frac{\Delta t}{12}\begin{bmatrix}
			0 & 0 & 0 & 0 & & \\
			5 & 8 & -1& 0&& \\
			0 & 5 & 8 & -1& & \\
			& \ddots & \ddots & \ddots & \ddots & \\
			& & 0 & 5& 8 & -1 \\
			& & 0 & -1& 8 & 5
		\end{bmatrix};
\end{equation*}
\begin{equation*}
	\bm{B}^{(3)} =  \frac{\Delta t}{24}\begin{bmatrix}
		0 & 0 & 0 & 0 & &\\
		9 & 19 & -5& 1&& \\
		-1 & 13 & 13& -1&& & \\
			& \ddots & \ddots & \ddots & \ddots & \\
			& & -1 & 13& 13 & -1 \\
			& & 1 & -5 & 19& 9
		\end{bmatrix}
\end{equation*}
\section{Data-preprocessing} \label{appC}
\subsection{Nonparametric denoising} When there is noise in the measurement, a unique solution to \cref{e2} is no longer guaranteed. The library of candidate functions are built from the noisy measurements, $\tilde{\bm{z}} = \bm{z} + \bm{\varepsilon}$, which are assumed to be corrupted by AWGN. With nonlinearities, noise can be nonlinear transformed when formed as a candidate function: e.g.~$\theta(\tilde{\bm{z}}_1\circ\tilde{\bm{z}}_2)=(\bm{z}_1^* + \bm{\varepsilon}_1)\circ(\bm{z}_2^* + \bm{\varepsilon}_2)$ which contains higher order noise corrupted terms. Data that is noise-free lets numerical derivatives, which are known to be unstable, have a lower error accumulation. This inspires \textit{a priori} denoising techniques for data-preprocessing before applying SINDy, e.g.~knowing the trajectory should be smooth. In this study, we opt for a global denoising techniques by regularization, which can be expressed as convex optimization. The most prominent in literature is by Tikhonov regularization, the right-hand side is penalized by an $\ell_2$ norm:
\begin{equation}
	\begin{aligned}
	\hat{\bm{z}} &= \argmin_{\bm{z}} \bigl \{\ \|\bm{z}-\tilde{\bm{z}}\|^2_2+\lambda\left\|\bm{D}_{2} \bm{z}\right\|^2_2 \ \bigr \}; \\
	&= \left(\bm{I}+\lambda \bm{D}_{2}^\mathrm{T} \bm{D}_{2}\right)^{-1} \tilde{\bm{z}};
\end{aligned}\quad  \forall \lambda > 0 \label{eb1}
\end{equation}
Note that denoising data with the penalization $\bm{D}_{2}$ corresponds to projecting its trajectory onto a smooth Poincar\'e map. If this is not suitable for the dynamics, e.g.~trajectories that are piece-wise smooth, one can use a combined Tikhonov and total-variation method \citep{gholamiBalancedCombinationTikhonov2013}.
It's been shown that this global smoothing technique, requiring one hyperparameter, outperforms various local smoothing techniques requiring multiple hyperparameters \citep{cortiellaPrioriDenoisingStrategies2022}.

\subsection{Regularized derivatives} The regression data $\dot{\bm{z}}$ is typically numerically estimated by finite difference approximations on the measurements $\bm{z}$. We opt for global methods by regularization, since it has been shown to outperform local methods \citep{knowles2014methods}, e.g.~finite differentiation on local windows which are smoothed by a Savitsky-Golay filter or LOWESS (moving average and/or polynomial fits).  For a $j$th order derivative, this can be written as the integration:
\begin{equation*}
	\bm{T}_j {\partial_j(\bm{z})} + z_0\bm{1}_M= \bm{z} + \bm{\varepsilon}_q 
\end{equation*}
where $\bm{\varepsilon}_q$, $z_0\bm{1}_M$ is the error due to quadrature and initial values respectively and $\partial_j(\bm{z})\coloneqq d^j/dt^j(\bm{z})$ notates the $j$th order derivative. This formulation has resulted into methods of differentiation to reduce the quadrature error and instabilities. Like previously in \cref{e10}, the expression can be $\ell_2$ regularized to approximate for the $j$th order derivatives $\partial_j(\bm{z}) \in \mathbb{R}^{M-j \times 1}$:
\begin{equation}
	\begin{aligned}
	\hat{\partial_j(\bm{z})} &= \argmin _{\partial_j(\bm{z})}  \bigl \{\ \|\bm{T}_j \partial_j(\bm{z})-{\bm{z}}\|_2^2+\lambda\|\bm{D}_2 \partial_j(\bm{z})\|_2^2 \ \bigr \}; \\
	&= \left(\bm{T}_j^\mathrm{T} \bm{T}_j+\lambda \bm{D}^\mathrm{T}_2 \bm{D}_2\right)^{-1}  \bm{T}_j^\mathrm{T} {\bm{z}};
\end{aligned}\quad  \forall \lambda > 0 \label{eb2}
\end{equation}
where $\bm{z}' = \bm{z} - z_0\bm{1}_M$. Here the finite difference matrix operators are used in the penalty, e.g.~$\bm{I}$, $\bm{D}_1$, $\bm{D}_2$ which penalize the amplitudes, the gradient, and the curvature of the solution respectively. Heuristically, $\bm{D}_2$ regularization balances the closeness to the data and the smoothness of the derivative. More importantly, the instability of finite difference is bypassed by reducing the numerical differentiation problem to a family of well-posed convex problems. These global method also allows for a Pareto front selection of the regularization parameter, where a standard-finite difference quadrature does not. The operator of $\bm{T}_1$ can be shown that it has an approximation error $\mathcal{O}((\Delta t)^2)$ (bias) and a noise amplification (variance) that scales $\mathcal{O}(\sigma^2/(\Delta t)^2)$. Finally, global methods tend to produce aliasing at the ends of the estimate, where the ends of the vector are trimmed.

\subsection{Scaling} Due to coefficient magnitude bias, some sparse regression estimators can be challenged to recover the true underlying dynamics if in two regards: one, if the estimator is LS-based, large condition numbers\footnote{The library basis has a condition number that is linear $\kappa(\bm{\Theta})$ or quadratic $\kappa(\bm{\Theta}^\mathrm{T}\bm{\Theta})$ depending on if the normal equations are used for an LS-based estimator.} will grossly degrade performance, i.e.~ill-conditioned; two, the regularization parameter effectively thresholds/shrinks the coefficients based on magnitude. Therefore, an $\ell_2$ norm scaling matrix $\bm{H}\in\mathbb{R}^{P\times P}$ is defined:
\begin{equation}
	\bm{H}_{ii} \coloneqq \sqrt{(\bm{\theta}_i^\mathrm{T}\bm{\theta}_i) },\quad \text{for}\ i = 1, \ldots, P\label{eb3}
\end{equation}
and when applied to SINDy framework of \cref{e3}:
\begin{equation}
	\dot{\bm{z}} = \breve{\bm{\Theta}}\breve{\bm{\xi}};\quad \text{where} \quad \breve{\bm{\Theta}} \coloneqq \bm{\Theta} \bm{H}^{-1}; \quad \breve{\bm{\xi}} \coloneqq \bm{H}\bm{\xi} \label{eb4}
\end{equation} 
Thus it's easy to see that there is a dual benefit: one, that  $\kappa(\breve{\bm{\Theta}})<\kappa(\bm{\Theta})$ and two, that the penalization of the scaled coefficients $\breve{\bm{\xi}}$ are equal weighted. Notably, scaling has been commonly practiced in statistical domains for Lasso-based regression \citep{harrellRegressionModelingStrategies2015} and in the PySINDy Python library \citep{kaptanogluPySINDyComprehensivePython2022a}, and is adopted in this study. 

\section{Simulation parameters for figures} \label{appB}
Both \cref{figure1} and \cref{figure2} were generated using the Bouc Wen system:
\begin{equation*}
	\left\{\begin{aligned}
	\dot{x}&=\sigma(y-x);  &\dot{x}(0) = -8\\
	\dot{y}&=\rho x-y-x z; &\dot{y}(0)=7 \\
	\dot{z}&=x y-\beta z; &\dot{z}(0)=27\\
	\end{aligned}\right.
	\end{equation*}
with the coefficients $\sigma = 10$, $\rho = 28$, $\beta = 8/3=2.6\bar{66}$. 

For \cref{figure1}, a four-second simulation of the Lorenz system using $\verb|ode45|$ with enforced time stepping of $\Delta t = 0.005$ yielded a signal of length $M=800$. The autocorrelative noise  with a level of $3\%$ was added to the path data using the function $R[n] = \sigma^2 * \exp(-|n| / 2)$. Tikhonov denoising was performed on the second DOF using \cref{eb1}, with L-curve and GCV criterions with the log-spaced hyperparameter grid  $\bm{\lambda}=\{10^-11,\ldots, 10^0\}$.

For \cref{figure2}, a three-second simulation of the Lorenz system using $\verb|ode45|$ with enforced time stepping of $\Delta t = 0.01$ yielded a signal of length $M=300$. AWGN noise with a level of $2\%$ s was added to both the path and trajectory data. Specifically, the third DOF was used for TRIM, IRL1 and STLS estimators: whose hyperparameters are summarized: STLS uses  log-spaced grid of size 100, $\bm{\varphi}=\{10^{-3},\ldots, 10^{3}\}$, for IRL1 a log-spaced grid of size 50 $\bm{\lambda}=\{10^{-20},\ldots, 10^{2}\}$ and ${q} = \{1\}$, and for TRIM $\bm{k}=\{1,2,\ldots, 13\}$ and $\nu=10$.

\section{Simultaneous estimation of Bouc Wen oscillator coefficients} \label{appD}
When the linear oscillator coefficients are to be identified simultaneously with the Bouc Wen's coefficients, we start from \cref{e23} with extra mass coefficient factored from the input:
\begin{equation*}
	\left\{\begin{array}{l}
	m\ddot{x} + c \dot{x} + k{x} + z = m u(t) \\[0.1cm]
	\dot{z}=\alpha\dot{x}-\beta |\dot{x}||z|^{\nu-1}z-\delta \dot{x}|z|^{\nu}
	\end{array}\right. 
	\end{equation*}
One can define the difference:
\begin{equation*}
	y \coloneqq  u(t) - \ddot{x} = \frac{c}{m} \dot{x} + \frac{k}{m}{x} + \frac{1}{m}z
\end{equation*}
With substitution of $z = my - c \dot{x} - k x$, this yields :
\begin{equation*}
	\left\{\begin{aligned}
	\ddot{x}&=y - u(t)\\
	\dot{y}&= \alpha\dot{x}- \beta |\dot{x}||z|^{\nu-1}z-\delta \dot{x}|z|^{\nu}\\
	&= \alpha\dot{x}- \beta |\dot{x}||my - c \dot{x} - k x|^{\nu-1}(my - c \dot{x} - k x)-\delta \dot{x}|my - c \dot{x} - k x|^{\nu}
	\end{aligned}\right.
	\end{equation*}
	Note that this form can be solved using a SINDy extension for hybrid systems \citep{thieleSystemIdentificationHysteresiscontrolled2020}. In \citep{laiSparseStructuralSystem2019}, the choice of $\nu = 1$ is adopted for ease of problem exposition, and we will further elaborate on this particular case. It is important to note that this selection of $\nu$ is not a constraint but rather a simplifying assumption for the purpose of clarity in the analysis. The absolute values yield two possible solutions, i.e.~$|my - c \dot{x} - k x| = \pm (my - c \dot{x} - k x)$. Therefore, one must solve for a piecewise problem, where we define $g \coloneqq my - c \dot{x} - k x$:
\begin{equation*}
	\left\{\begin{aligned}
	\ddot{x}&=y - u(t)\\
	\dot{y}&= \left\{\begin{aligned}
		\alpha\dot{x}-\beta |\dot{x}| (my - c \dot{x} - k x)-\delta \dot{x} (my - c \dot{x} - k x) & \quad \text {if } g \geq 0 \\
		\alpha\dot{x}-\beta|\dot{x}| ( c \dot{x} + k x - my)-\delta \dot{x} ( c \dot{x} + k x - my) & \quad \text {if } g<0
		\end{aligned}\right.
	\end{aligned}\right.
	\end{equation*}
Expanding this out and collecting like terms yields:
\begin{equation}
	\left\{\begin{aligned}
	\ddot{x}&=y - u(t)\\
	\dot{y}&= \left\{\begin{aligned}
		\alpha \dot{x}-\beta m |\dot{x}|y+\beta c |\dot{x}|\dot{x}+\beta k |\dot{x}|x-\delta m \dot{x}  y+\delta c  \dot{x}^2+\delta  k \dot{x}x &\quad \text {if } g \geq 0 \\
		\alpha \dot{x}+\beta m |\dot{x}|y-\beta c |\dot{x}|\dot{x}-\beta k |\dot{x}|x+\delta m \dot{x}  y-\delta c  \dot{x}^2-\delta  k \dot{x}x &\quad \text {if } g<0
		\end{aligned}\right.
	\end{aligned}\right.\label{de1}
	\end{equation}
Here, $g$ can be approximated to give the inflection points, i.e.~knots, using  regression on: $\check{g} \coloneqq \check{m}y - \check{c} \dot{x} - \check{k} x$. Thus approximated knots $g \leftarrow \check{g}$ are used and the sparse regression can be performed for \cref{de1}. Note, the library matrix must be constructed with the minimum of these candidate functions for the second DOF state as shown in \cref{de1}.

\bibliographystyle{ieeetr}
\bibliography{s-SINDy.bib} 
\end{document}